\documentclass{article}

\usepackage[final]{corl_2019} % Uncomment for the camera-ready ``final'' version
\usepackage{amsmath}
\usepackage{graphicx,psfrag,picture}
\usepackage{subfigure, xcolor}

\title{Energy-efficient Path Planning for Ground Robots by Combining Air and Ground Measurements}

\author{
  Minghan Wei And Volkan Isler\\
  Department of Computer Science and Engineering\\
  University of Minnesota 
  United States\\
  \texttt{\{weixx526, isler\}@umn.edu} \\
}

\begin{document}
\maketitle

%===============================================================================
\begin{abstract}
As mobile robots find increasing use in outdoor applications, designing energy-efficient robot navigation algorithms is gaining importance. There are two primary approaches to energy efficient navigation: 
Offline approaches rely on a previously built energy map as input to a path planner. Obtaining energy maps for large environments is challenging. Alternatively, the robot can navigate in an online fashion and build the map as it navigates. Online navigation in unknown environments with only local information is still a challenging research problem. In this paper, we present a novel approach which addresses both of these challenges. Our approach starts with a segmented aerial image of the environment. We show that a coarse energy map can be built from the segmentation. However, the absolute energy value for a specific terrain type (e.g. grass) can vary across environments. Therefore, rather than using this energy map directly, we use it to build the covariance function for a Gaussian Process (GP) based representation of the environment. In the online phase, energy measurements collected during navigation are used for estimating energy profiles across the environment using GP regression. Coupled with an $A^\star$-like navigation algorithm, we show in simulations that our approach outperforms representative baseline approaches. We also present results from field experiments which demonstrate the practical applicability of our method.
\end{abstract}

% Two or three meaningful keywords should be added here

%===============================================================================
\section{Introduction}
%Many applications need ground robots operate in the fields, such as environmental monitoring, mapping, and search. 
Autonomous navigation in outdoor fields is a crucial capability for a wide range of robotics applications such as environmental monitoring, agriculture, search and mapping. In these applications, efficiency is a major concern when using energy constrained robots in large environments. The energy profiles of most outdoor fields can be rather complex due to varying surface properties (grass, dirt, mud, etc.) and geometry. Computing energy efficient paths in such environments remains an important robotics problem.

\begin{figure}
\centering
\includegraphics[width=0.8\columnwidth]{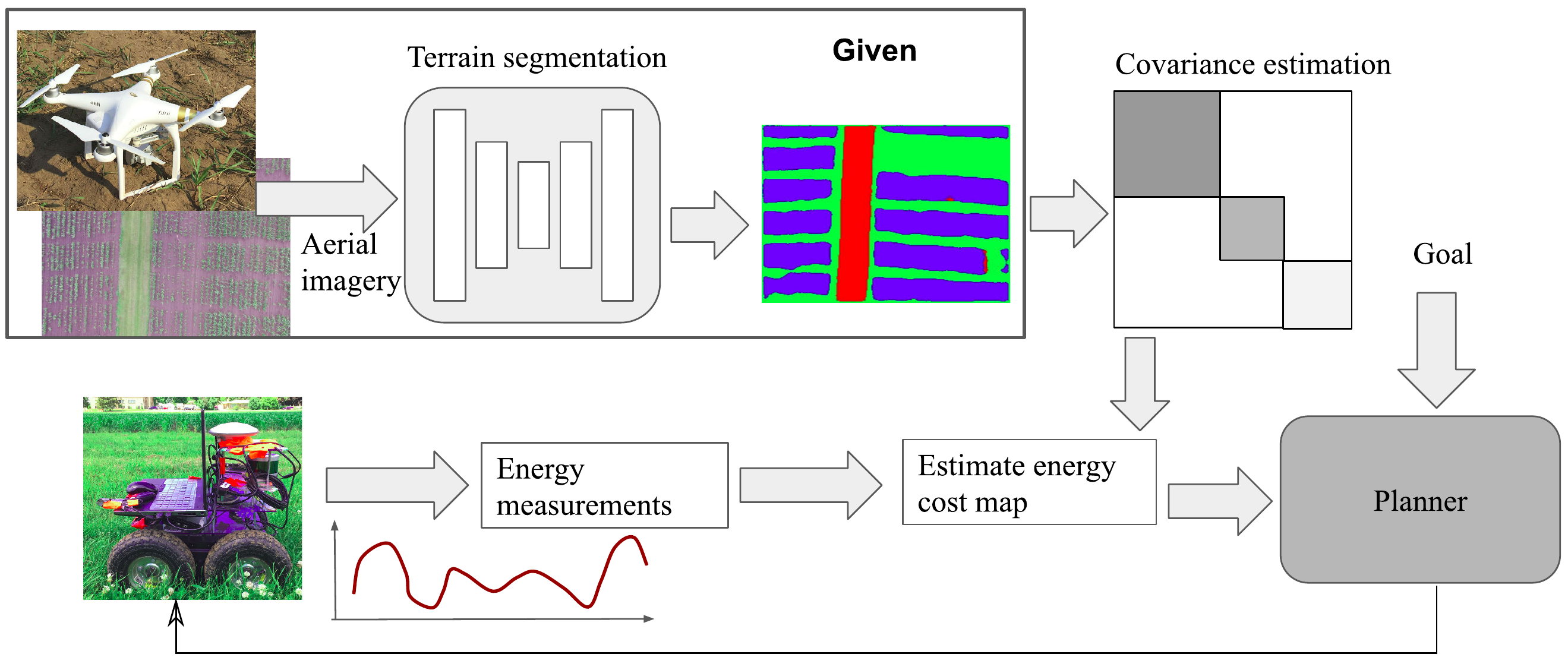}
\caption{An overview of our approach. Top row: Aerial images with the segmentation are used to build the covariance function for GP regression for the energy cost map. Bottom row: during navigation, online measurements along with the estimated covariance matrix are used to build an energy map for path planning.
\label{fig:concept}}
\end{figure}

Path planning, including variations focusing on energy efficiency, has been widely studied. Some of these algorithms plan energy-efficient paths offline, where an energy-cost map of the field is used as the input~\cite{sun2005finding}. These methods can find the optimal path, but it is not easy to have an energy cost map in advance for path planning, especially for large fields. There are online algorithms that focus on exploring the environment and building a map~\cite{salan2014minimum}. It may be unnecessary to explore the entire area using the ground robot just for planning a path. Robots that navigate online with only local measurements need to replan the paths \cite{ersson2001path}, thus the online costs can be much higher than the offline solutions. 

We study the scenario where the ground robot navigates in a previously unknown field whose aerial images and terrain classes are available. Such images can be obtained from aerial vehicles such as drones or satellites. The terrain classes can be obtained using existing segmentation methods. Fig.~\ref{fig:concept} shows an overview of our approach which starts with the segmented images. From the segmentation we know the terrain classes of each location. However they are not directly used to estimate an energy cost map, since the absolute values of energy consumption on the same terrain class can vary even for the same ground robot. For example, we measured the energy cost on grass at different sites. The average cost can vary from $50\sim70J/m$ measured with the ground robot in Fig.~\ref{fig:concept}. As a reference, the concrete road at one measured site takes $40J/m$ on average. The variation on grass is almost $50\%$ of the road, which will significantly affect path planning. Therefore, we build a joint Gaussian distribution of the energy cost over the environment. The covariance is initialized using terrain class information. The energy map is then learned by GP regression using the onboard energy cost measurements during  navigation. The main contributions of this paper are: $(1)$~We present a Gaussian Process model of combining terrain appearance and robot onboard measurements for energy-efficient ground robot navigation. $(2)$~We test this model via simulations and experiments. The results show that the planned paths by this model consume less energy compared to shortest-path planning algorithms and GP-based approaches which only use distance-based covariance functions. 
%===============================================================================
\section{Related Work}
When the cost is uniform across the environment, planning an energy efficient path becomes equivalent to finding the shortest path. In this case, Dijkstra's algorithm, or heuristic search methods such as $A^*$ can be used to find the optimal path. To improve the planning efficiency, RRT algorithm~\cite{lavalle1998rapidly}, and its variants \cite{kuffner2000rrt} \cite{atramentov2002efficient} are also used in the literature. When the robot navigates in unknown environment, $D^*$ algorithm is useful for fast planning \cite{koenig2005fast}.  

A field robot may have to navigate in rough environments. The work in \cite{howard2007optimal} presented an efficient method to generate trajectories for arbitrary terrain shape by considering the dynamics of a practical vehicle. We refer readers to survey \cite{paden2016survey} for additional path planning  and robot controlling methods .

However, the energy consumption during navigation is affected by the environment properties as well as the robot's own motion profile. Thus minimizing path length is not sufficient. Some prior work uses environment properties to improve path efficiency. The energy consumption of the ground robot can be modeled based on  physical laws such as $E = mg(\mu \cos{\theta} + \sin{\theta})l$, where $\mu$ is the friction coefficient, $\theta$ is the inclination angle, and $l$ is the moving distance \cite{tiwari2018estimating} \cite{sun2005finding} \cite{liu2014minimizing}. Constraints such as impermissible heading due to the ground inclination can also be added \cite{ganganath2015constraint}. With this model we can build an energy cost map of the environment and then plan an optimal path. But obtaining the environment properties such as friction coefficients is not easy over large environments. Also, the energy cost of the jerky motion on rough terrain surfaces can not be easily modeled by physics laws. Our method addresses these issues using a data-driven approach. We model the energy consumption stochastically using aerial images to estimate the energy cost correlations across the terrain. 

Visual inputs have recently been used for ground robot navigation. Ground traversability can be learned from images \cite{schilling2017geometric,chavez2018learning}. Delmerico et al. planned paths for the UAV and the ground robot simultaneously so that the sum of the exploration (by the UAV) and navigation time (by the ground robot) can be minimized. These papers do not consider the energy cost of the planned paths.

Finding an optimal velocity profile for a given path can also improve the energy efficiency \cite{mei2004energy} \cite{tokekar2014energy}, since the acceleration and turning usually costs more energy than moving in constant velocity. In this paper, we use onboard measurements of the ground robot so that the navigation and energy-cost map building can be done simultaneously. 

Our work is also related to the topic of efficient exploration of the environment. Some work view the exploration as an online coverage path planning problem \cite{mei2006energy} \cite{salan2014minimum}, where the goal is to explore the whole area with minimum cost. Our work differs from these in that it is not necessary to cover the whole environment just for navigating to a position. Related work aims to maximize the information gain within a cost \cite{chekuri2005recursive} \cite{singh2009efficient}, while our goal is to reach a position and minimize the cost. 

\section{Problem Formulation}
\label{sec:formulation}
We consider the case of a robot moving at constant forward speed. The goal is to move from a starting location $x_0$ to a goal location $x_g$ in such a way that the total energy consumption along the trajectory is minimized. We now present details of the energy cost model.

Let $e(x)$ be the energy consumption for moving a unit distance at a position $x$. For a given trajectory, the energy consumption is given by the product of the motor voltage and current integrated along the trajectory:
\begin{equation}
E = \int_{t_0}^{t_1} V(t)I(t)dt = \int_{x_0}^{x_1} e(x)dx 
\label{eqn:energy}    
\end{equation}
where the second integral is taken along the trajectory and the constant velocity assumption is used to substitute the integral variable.

We assume that terrain classes of the environment are given. Let $P_x(e)$ be the probability that energy consumption at location $x$ is $e$. We model the distribution of $P_x$ as a Gaussian whose mean and variance depends on the terrain type $T$ of $x$: $P_x(e) = \mathcal{N}(\mu_T, \sigma_T)$. We drop the subscript $x$ to simplify the notation.

Now consider two locations $x_1$ and $x_2$, and the unit-distance energy consumption $e_1$ and $e_2$ at these locations. Let $c(x)$ denote the terrain type of $x$. There are two cases:
$\mathbf{(i)}$~$c(x_1)=c(x_2)$. In this case, the joint distribution is given by 
$P(e_1, e_2) = \mathcal{N}(\mathbf{\mu_T}, \mathbf{\sigma_T})$,
where $\mathbf{\mu_T}=(\mu_1, \mu_2)$ and 
$\mathbf{\sigma_T}$ is a symmetric $2 \times 2$ matrix with diagonal terms corresponding to the estimation uncertainty. We use the squared exponential function for the off-diagonal terms (covariance): 
$\mathbf{\sigma_T}(1,2) = \mathbf{\sigma_T}(2,1) = {\sigma^2}_f \exp(-\frac{1}{2}\frac{ (x_i - x_j)'(x_i - x_j)}{\sigma_d^2})$. Thus $x_1$ and $x_2$ become more correlated as the distance between them gets smaller. 
$\mathbf{(ii)}$~$x_1$ and $x_2$ have different terrain types. We assume that the energy consumption at these two locations are independent:  $P(e_1, e_2) = P(e_1)P(e_2)$.

Building on the cases $(i)$ and $(ii)$, we model the energy consumption over the entire environment as a spatial Gaussian Process (GP). The joint distribution is $\mathcal{N}(\mathbf{u}, \mathbf{\sigma})$, where $\mathbf{u}=\{u_1, u_2, ..., u_n\}$ is the mean energy consumption at locations $X =\{x_1, x_2, ..., x_n\}$ in the environment, and the kernel function (covariance) is
\begin{equation}
\sigma(x_i, x_j)^2 = 
\begin{cases}
0, & c(x_i) \neq c(x_j) \\
{\sigma^2}_{i,j,f} e^{-\frac{1}{2}\frac{ (x_i - x_j)'(x_i - x_j)}{\sigma_{i,j,d}^2}}, & c(x_i) = c(x_j)
\end{cases}
\label{eq:joint}
\end{equation}
The kernel function uses the terrain segmentation from the images. Note that in GP model the prior mean $\mathbf{u}$ can be set to zero for simplifying the computation. Setting the prior to $0$ is achieved by subtracting the prior mean from all observations.

The problem formulation as presented assumes environments with zero ground inclination angle. We can address terrain geometry in a way similar to the work in \cite{tiwari2018estimating} \cite{sun2005finding} by adding an additional cost of 
\begin{equation}
G = mg\sin{\theta}l,
\label{eq:gravity}
\end{equation}
when measuring the unit-distance energy cost at a location using Eqn.~\ref{eqn:energy}, where $mg$ is the weight of the robot, $\theta$ is the inclination angle, and $l$ is the moving distance. The ground inclination angle information can be obtained from a UAV equipped with a Lidar, or 3D reconstruction using the aerial images of the field. For convenience, the rest of the paper will focus on environments with zero inclination angle since the gravity can be incorporated if slope information is available.

To summarize, given an environment $X$, a terrain type for each location $x \in X$, an initial mean and covariance for each terrain type as well as the starting and goal locations of the robots, we seek a trajectory which connects these two locations and  minimizes the energy consumption along the way given by Eqn.~\ref{eqn:energy}.

%In Sec.~\ref{sec:classification} we describe how we obtain terrain classes from aerial images. Afterwards, 
We present the solution to this optimization problem in Sec.~\ref{sec:navigation}. Then in Sec.~\ref{sec:result} we compare our method to baseline methods. Here, we also present how we obtain terrain classes from aerial images (Subsection.~\ref{sec:exclassification}).

\section{Method}
\label{sec:navigation}
In this section, we present our algorithm for navigating the robot to the goal position in an energy efficient manner. Our method combines the terrain class information with onboard measurements to estimate an energy-cost map, which is used as input to a path planner. The planner chooses the next location where new energy measurements become available. 

For simplicity, we use a uniform grid to represent the energy cost map of the environment. We first partition the environment into uniform cells and build a dual graph whose vertices correspond to grid cells. Edges connect the vertices of adjacent cells. We assume that the unit-distance energy cost $e$ is uniform within a cell. To measure $e$ at a cell during navigation, we integrate the energy cost using Eqn.~\ref{eqn:energy}. Let $c_1, c_2$ be two adjacent cells, and $e_1, e_2$ be the unit-distance energy cost in $c_1, c_2$. The cost (weight) of the edge between $c_1, c_2$ is $E_{c_1,c_2} = (\frac{e_1+e_2}{2}) d(c_1, c_2)$, where $d(c_1, c_2)$ is the distance between the centers of $c_1$ and $c_2$. With this graph representation, we can use any path planning algorithm to plan an energy-efficient path from the robot current position to the goal. In this paper, we use Dijkstra's algorithm.

The weights in the energy cost map are initialized with available prior knowledge, which can be obtained from past experiences when the robot navigates in other environments. In Sec.~\ref{sec:explore}, we discuss the initialization strategy.

Our algorithm navigates the robot to the goal position while updating the energy cost map. The robot computes a path based on the current estimation of the map and follows the path to visit $m$ new grid cells. Then the map is updated based on the newly collected energy data at the $m$ cells. The parameter $m$ corresponds to the desired update frequency. The robot continues the navigation using the updated energy cost map. This process is repeated until the robot reaches the goal. 

We update the map as follows. The new data is $M = \{(x_1, y_1, e_1), (x_2, y_2, e_2), ..., (x_m, y_m, e_m)\}$, where $(x, y)$ is the position of the measurements and $e$ is the energy cost values. Together with all the measurements since the starting position, we update the map using the standard results from~\cite{rasmussen2010gaussian}.
\begin{equation}
\begin{aligned}
           \mu & = K(X_{*}, X_t)[K(X_t, X_t)+\sigma_{\epsilon}^2 I]^{-1} e_t\\
  K(X_{*},X_*) & = K(X_{*}, X_{*}) - K(X_*, X_t)[K(X_t, X_t)+\sigma_{\epsilon}I]^{-1}K(X_t,X_{*})  \\
\end{aligned}
\end{equation}
where $X_t$ is set of the measurements, $X_*$ is the set of positions to predict, and $K$ is the kernel function (covariance) as defined in Eqn.~\ref{eq:joint}.

\subsection{Initialization and Exploration of Unmeasured Terrain Classes}
\label{sec:explore}
By Eqn~\ref{eq:joint}, the unit-distance energy costs at two locations of different classes are uncorrelated. Therefore, the Gaussian Process model cannot update the energy cost data for the areas of unmeasured classes. Here an {\em unmeasured class} means no measured locations belong to this class. {\em Exploring an unmeasured class} means navigating to the closest location of this class in the map to take energy cost measurements. The path planner has to use prior knowledge to compute a path. Therefore, initialization is important for finding an energy-efficient path in our method.

\begin{figure}
\centering
\subfigure[]{
	\includegraphics[width=.22\columnwidth]{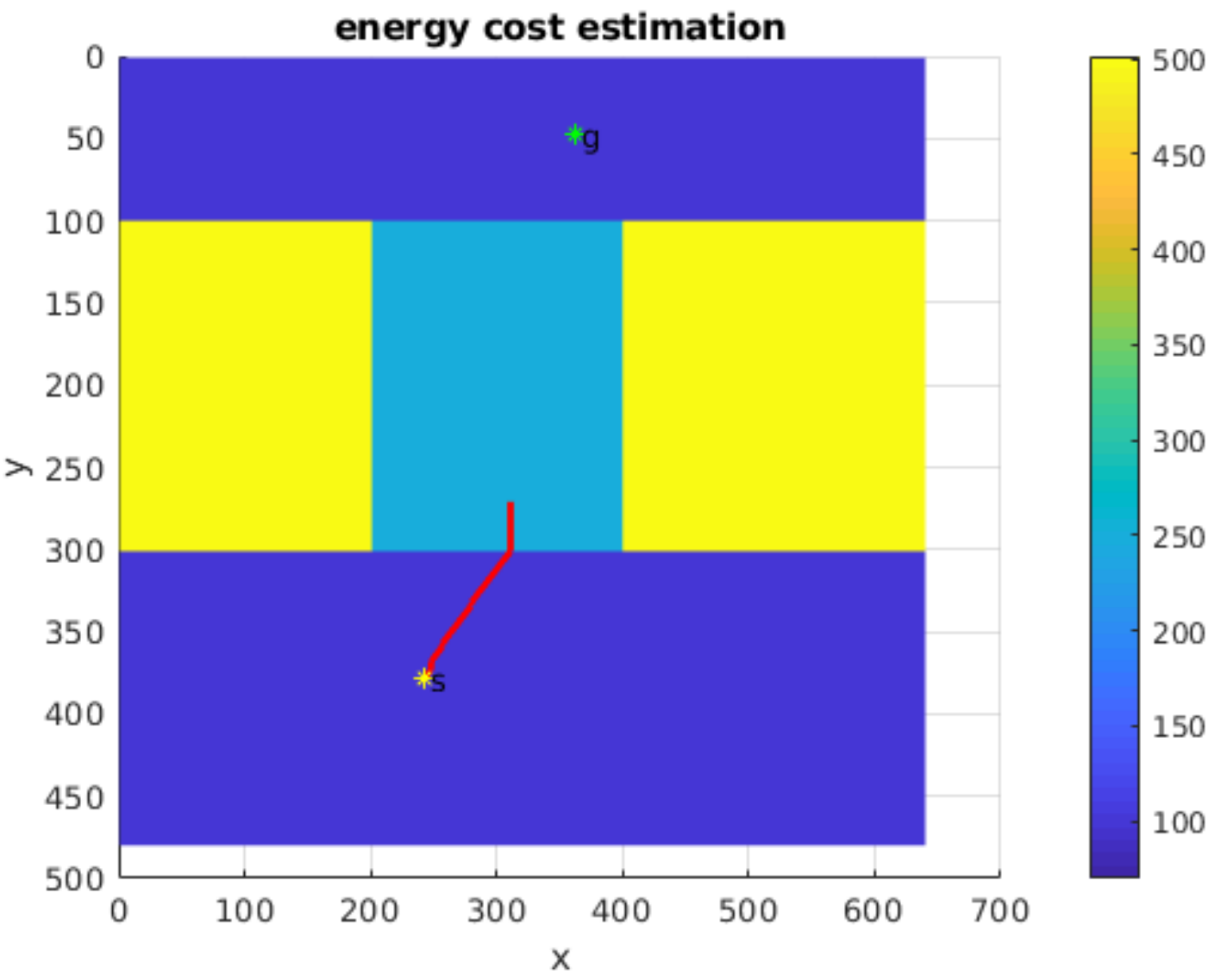}
	\label{fig:state2high}
}
\subfigure[]{	
	\includegraphics[width=.22\columnwidth]{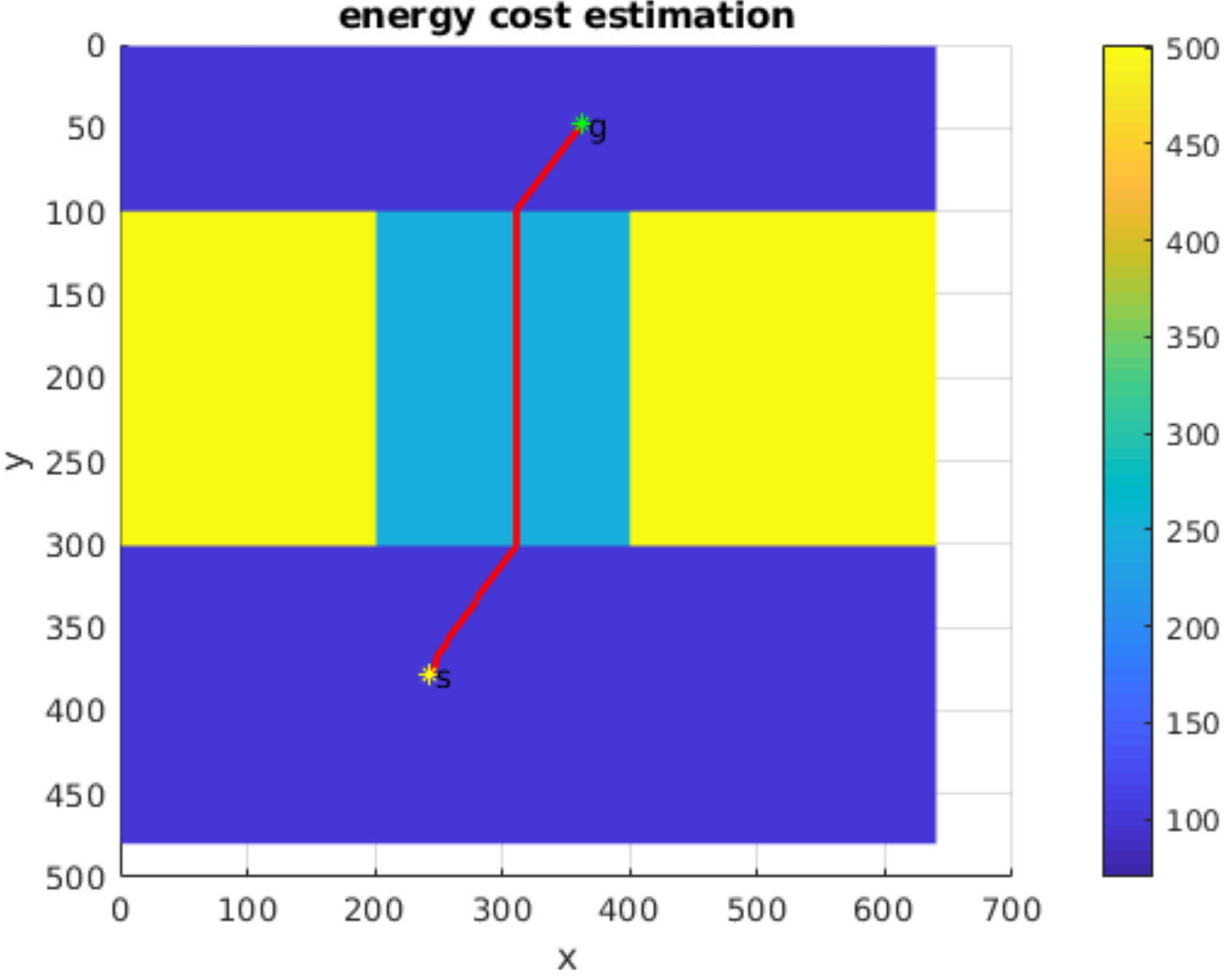}
	\label{fig:state3high}
}
\subfigure[]{
	\includegraphics[width=.22\columnwidth]{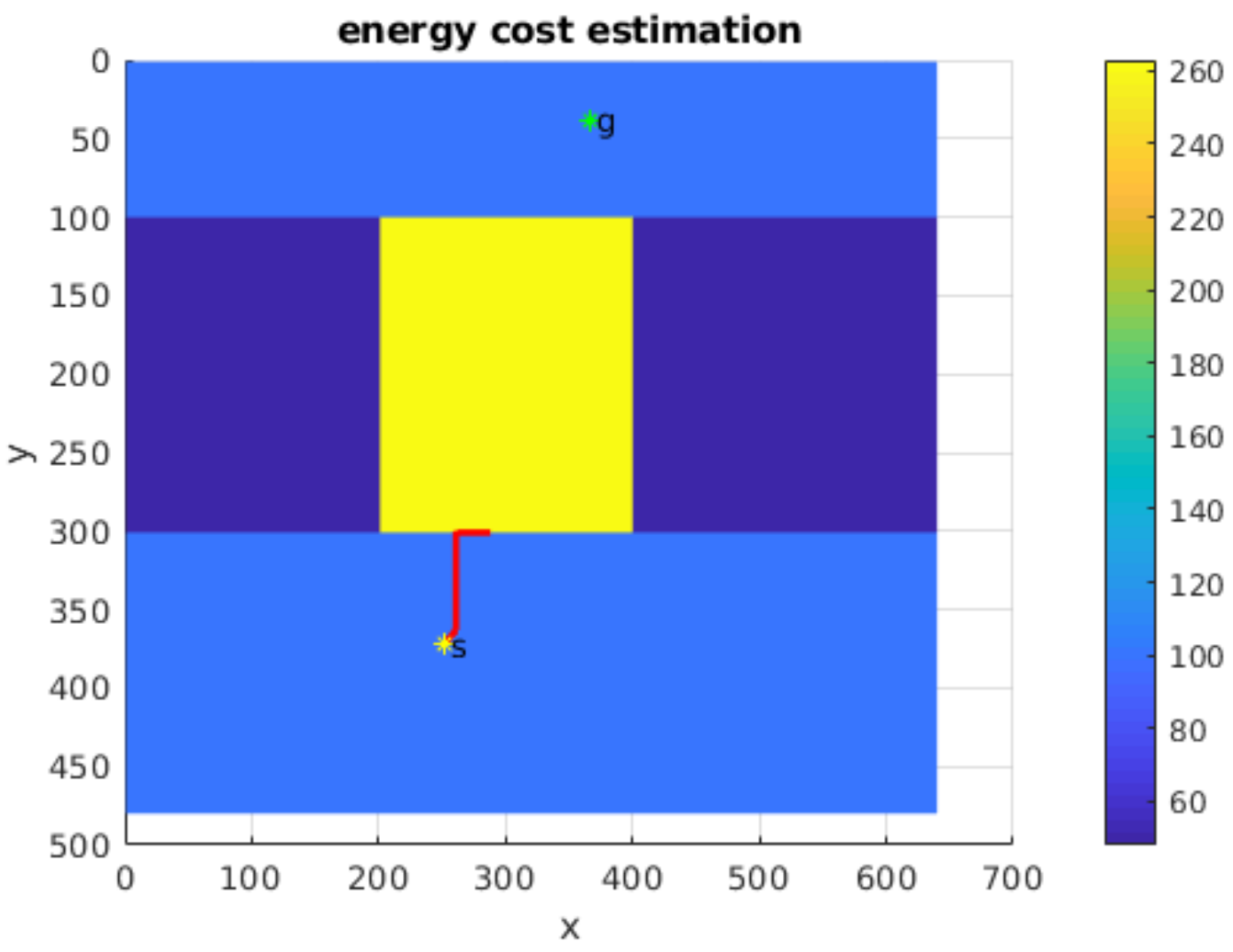}
	\label{fig:state2low}
}
\subfigure[]{	
	\includegraphics[width=.22\columnwidth]{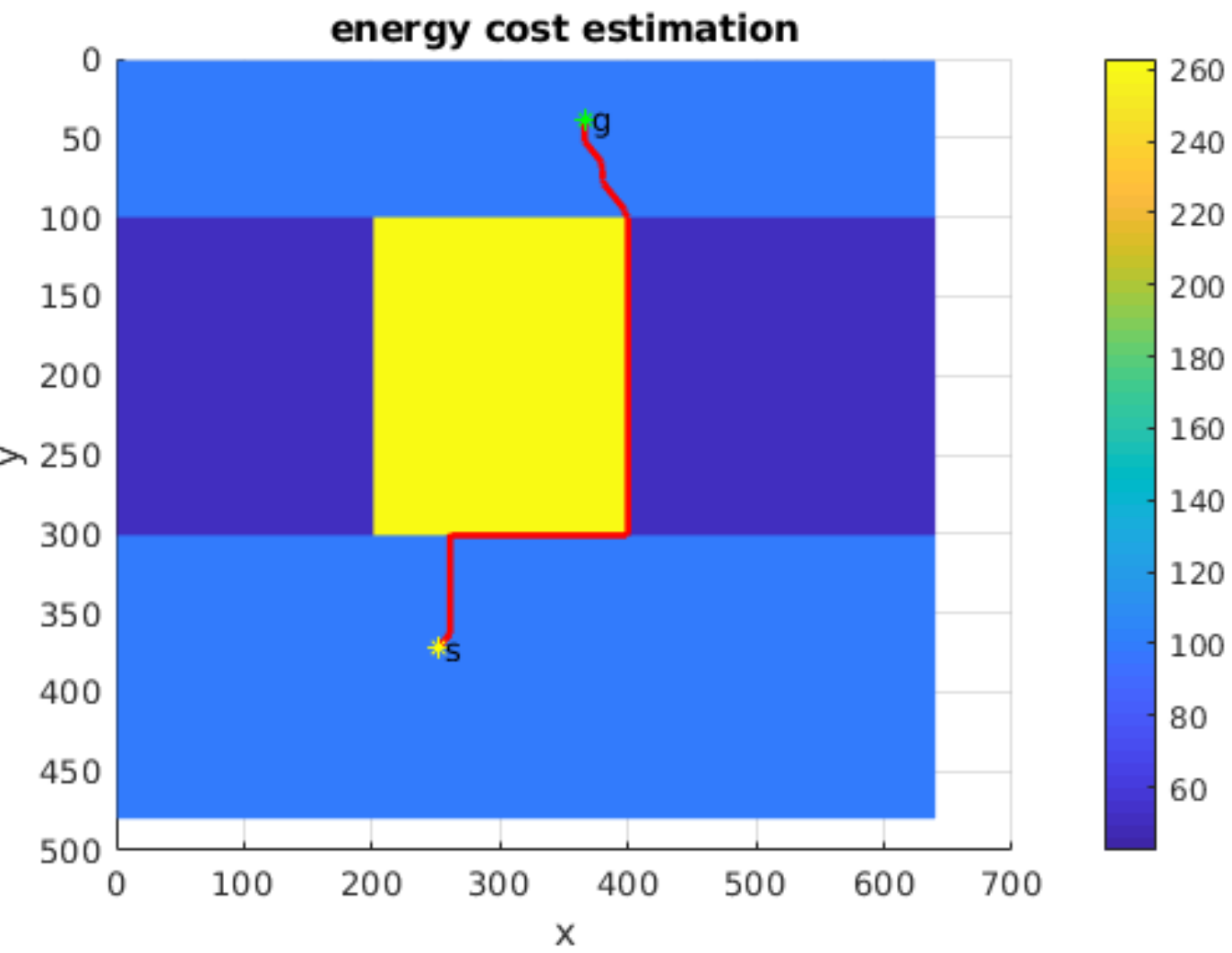}
	\label{fig:state3low}  
}
\caption{The robot updates the map during navigation. The red lines are the robot trajectories. $(a),~(c)$~The robot's trajectory and the map estimation when the robot has measurements on class $1$ and $3$. $(b)$~The robot navigates to the goal without exploring the unmeasured classes (class $2$) when the initial cost is high. $(d)$~The robot explores an unmeasured class in $(b)$ (class $2$) and finds a more efficient path with the admissible initialization.}
\label{fig:step1}
\end{figure}

Our initialization strategy yields an admissible heuristic for $A^*$. Specifically, we initialize the energy cost for unmeasured classes with a small value that does not overestimate their actual costs. In practice, this cost can be the smallest unit-distance energy cost measured in the past navigation in other environments. We identify two advantages of this admissible heuristic in our application.
$\textbf{(i)}$~If we do not use the admissible heuristics and initialize the unmeasured classes with a large cost instead, the robot does not explore the unmeasured classes even if they can lead to a more efficient path, as shown in Fig.~\ref{fig:state3high}.
$\textbf{(ii)}$~With this initialization, the algorithm can automatically trade off between navigating to the goal and exploring the unmeasured classes. More specifically, when an unmeasured class is initialized with a low cost but it takes too much energy to explore it, the path planner will not explore this class. On the other hand, if the exploration cost plus the cost to navigate to the goal via this unmeasured class is optimal based on the current energy cost map, the algorithm will return a path that explores this class. In this way, the admissible initialization balances the navigation and exploration costs.

\begin{figure}
\centering
\subfigure[]{
	\includegraphics[width=.22\columnwidth]{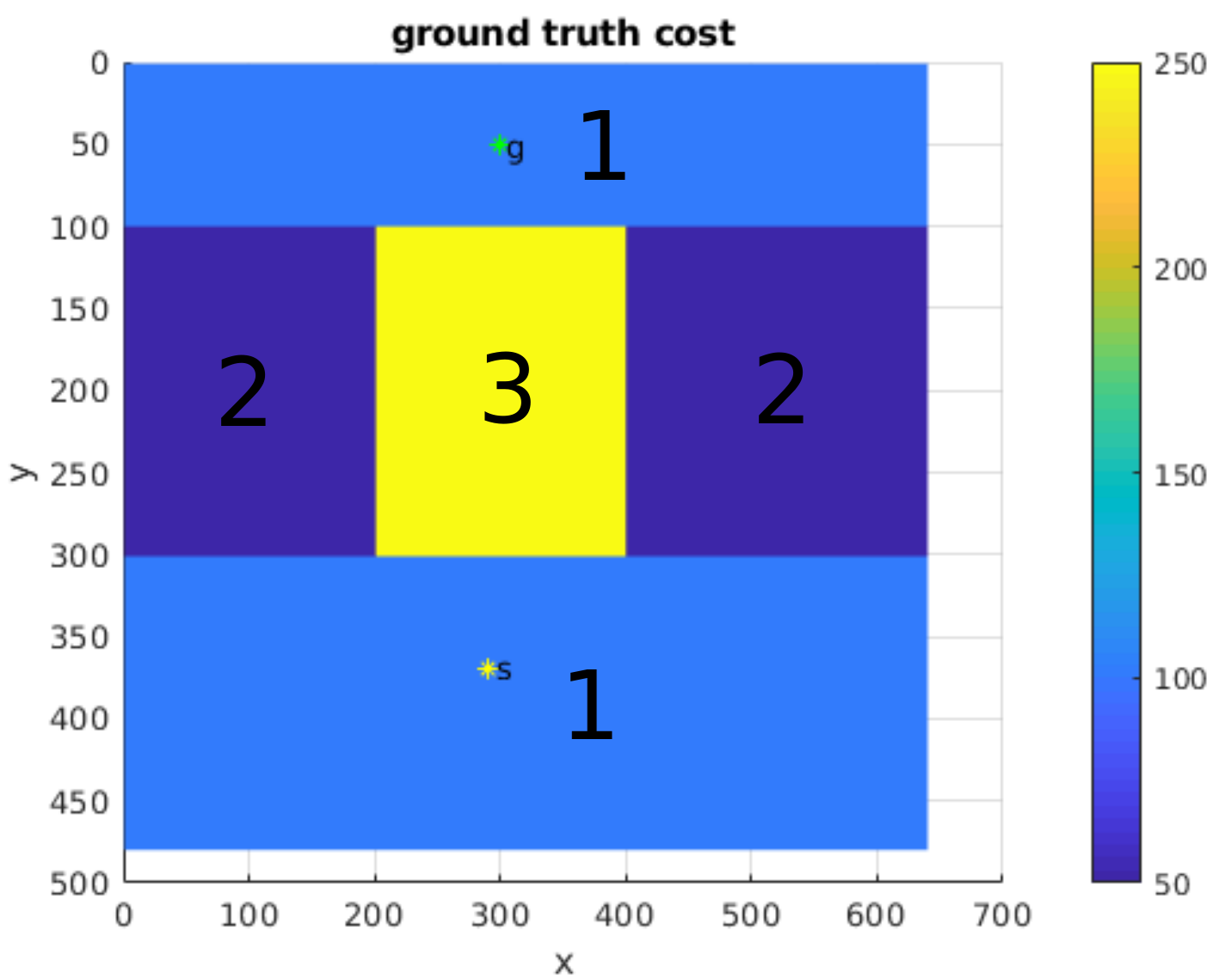}
	\label{fig:synenv}
}
\subfigure[]{	
	\includegraphics[width=.24\columnwidth]{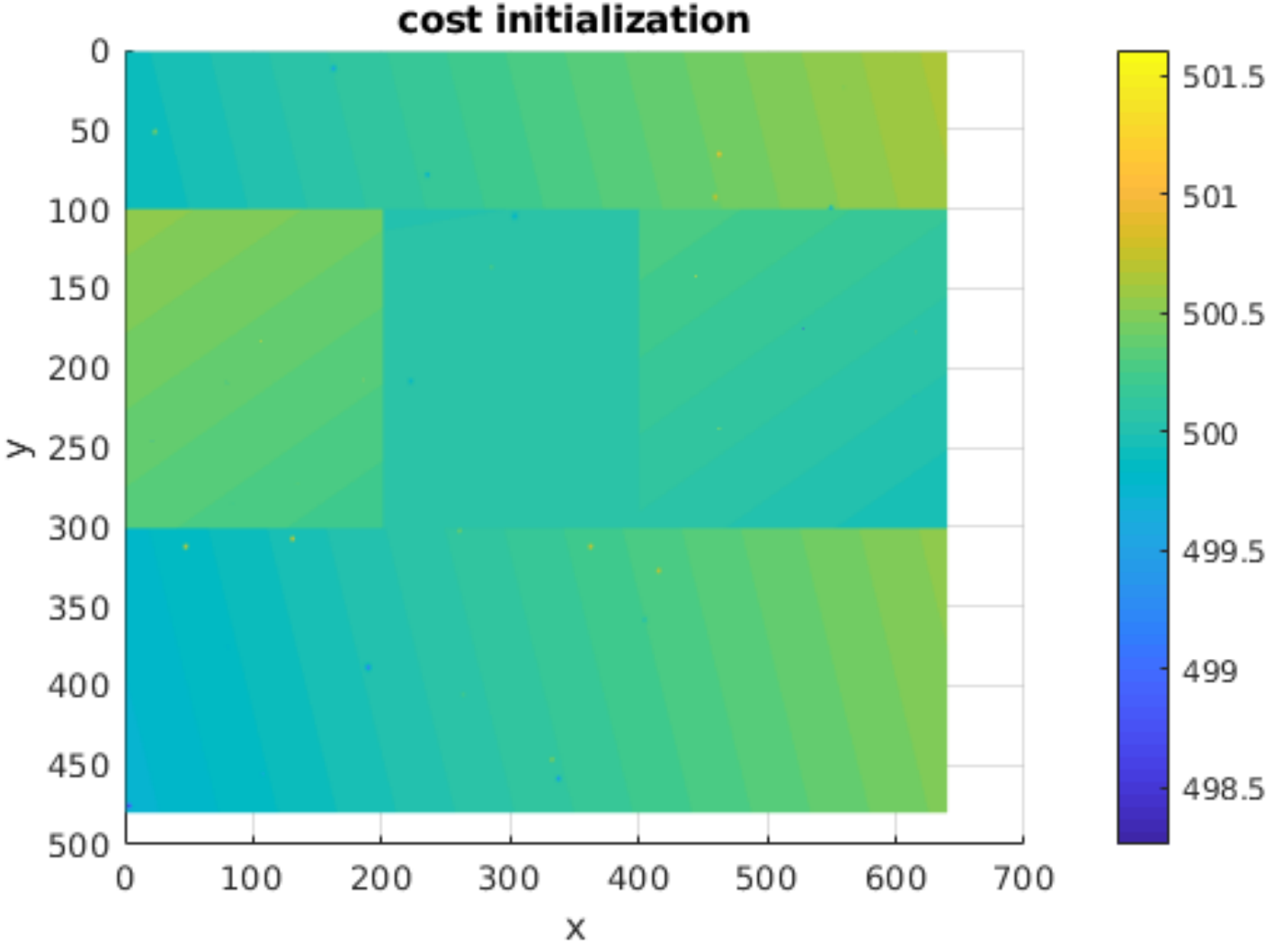}
	\label{fig:inicost1}  
}
\subfigure[]{
	\includegraphics[width=.22\columnwidth]{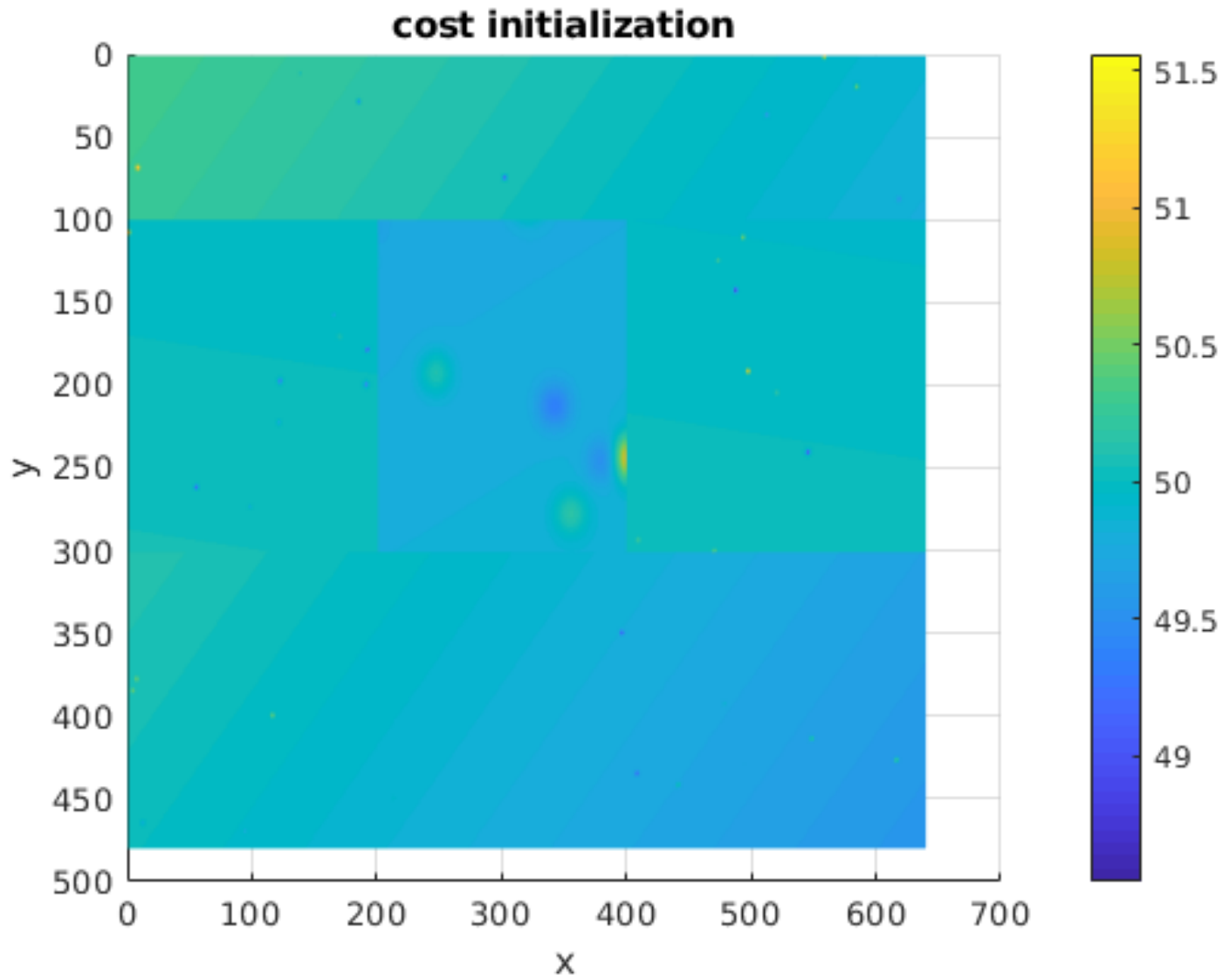}
	\label{fig:inicost2}
}
\caption{The environment the robot navigates in. The colors reflect the energy cost at the corresponding position. To show the effect of the initialization, we initialize the map with different values for navigation. $(a)$ The ground truth for the unit-distance energy cost over the environment, which is unknown to the robot in the beginning. $(b)~(c)$ The energy cost map is initialized with a high and low cost, respectively.}
\label{fig:env}
\end{figure}

We use the following synthetic example to illustrate the algorithm. Fig.~\ref{fig:synenv} shows the environment the robot navigates in. It consists of three terrain classes. The color bar shows the corresponding unit-distance energy cost values at each position. In the beginning, the robot does not have precise prior energy cost information. We compare two cases where in one case we initialize the whole environment with a low cost, and a high cost in the other. Fig.~\ref{fig:inicost1} and Fig.~\ref{fig:inicost2} show the energy values corresponding to the two initialization schemes.

The robot plans a path according to the current estimation of the map. During navigation, the robot gets new measurements to update the energy cost map. In this example, the map is updated every ten pixels ($m=10$). In Fig.~\ref{fig:state2high} and Fig.~\ref{fig:state2low}, the robot gets information about the terrain class $``1"$ and $``3"$. The class $``2"$ remains the same as the initial value. 
The initialization values affect whether the robot explores the unmeasured classes or not. When the initial cost is high, the robot navigates directly to the goal without exploring the unmeasured class, as shown in Fig~\ref{fig:state3high}. When the initial cost is low, the robot finds a more efficient path by exploring the previous unmeasured class (class $2$ in this example), as shown in Fig.~\ref{fig:state3low}.

\section{Experimental Results}
\label{sec:result}
In this subsection, we first introduce how we obtain the aerial images and terrain classes, which is assumed to be known in as mentioned in Sec.~\ref{sec:formulation}. Then we compare our results to baseline methods in simulations and experiments.

\subsection{Data Collection and Terrain Classification}
\label{sec:exclassification}
We collected aerial images at a corn field located in Saint Paul, Minnesota with a DJI Phantom 3 Professional UAV. The field includes three terrain types: corn, grass, and dirt road. 

To obtain the terrain classes, we segment the aerial images using u-net~\cite{ronneberger2015u}. The network takes RGB images as input and outputs the masks of the same size as the input images, where each element in the mask represents the class label of the corresponding pixel.

\begin{figure}[ht]
\centering
\subfigure[]{
	\includegraphics[width=.145\columnwidth]{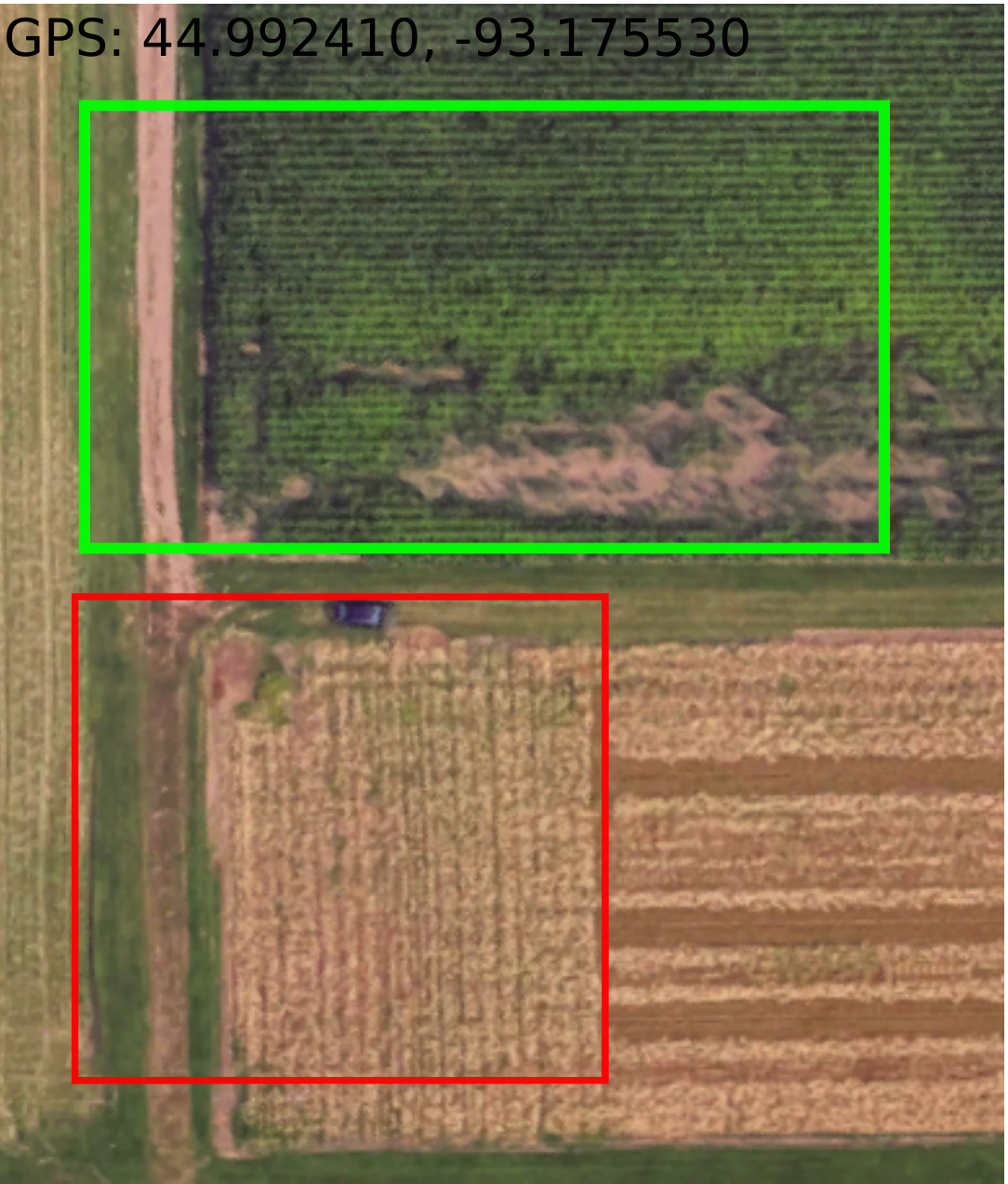}
	\label{fig:exmap}
}
\subfigure[]{
	\includegraphics[width=.225\columnwidth]{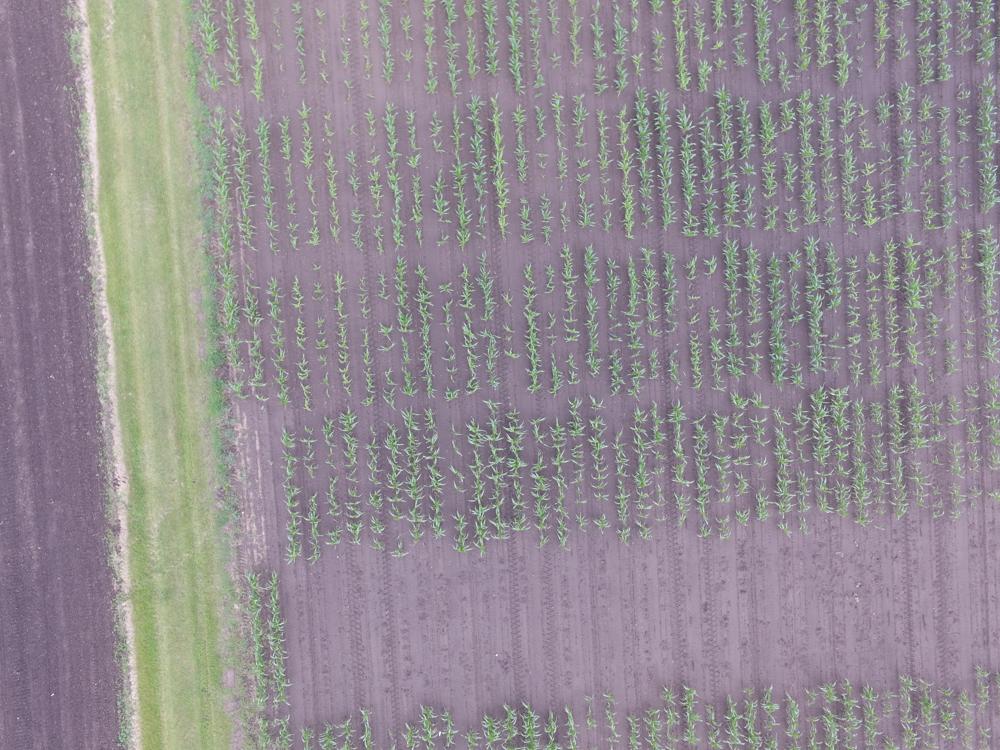}
	\label{fig:corntrain}
}
\subfigure[]{	
	\includegraphics[width=.22\columnwidth]{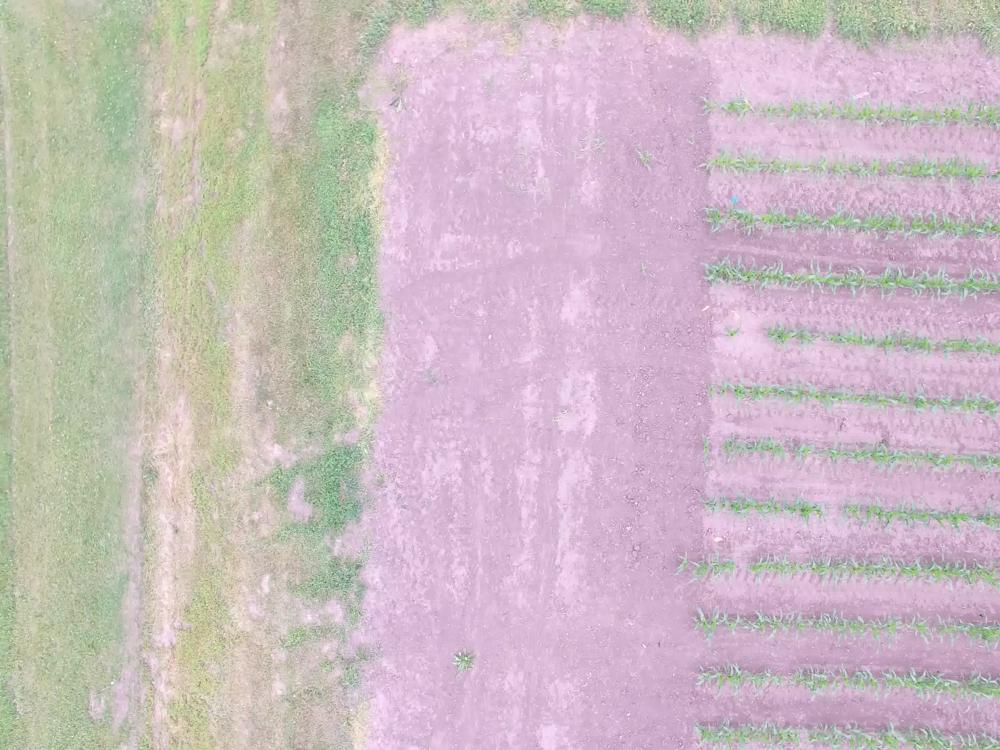}
	\label{fig:contest}
}
\subfigure[]{
	\includegraphics[width=.16\columnwidth]{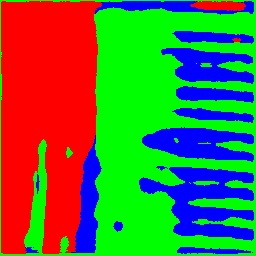}
	\label{fig:test1}
}
\subfigure[]{	
	\includegraphics[width=.16\columnwidth]{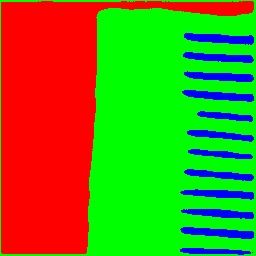}
	\label{fig:test1old}  
}
\caption{The terrain images and the classification result. $(a)$~In the training data, corn rows were planted close to each other. Thus there were no dirt roads between two rows. $(b)$~The corn rows were more sparsely planted in the testing set. $(c)$ The network by training images only does not give good performance due to the difference between training and testing images. $(d)$ After retraining, the segmentation result is improved.}
\label{fig:fielddiff}
\end{figure}

The dataset for training the network (training set) was collected on June 29, 2017. The green rectangle in Fig.~\ref{fig:exmap} marks the area. It contains 253 images in total and covers an area of $1500m^2$. To train for terrain classification, we labelled 97 images and use half of the labelled images for training and the other half for validation. The images are scaled to $256 \times 256$. We train the network on a Dell Precision 7530 machine with a Nvidia Quadro P3200 GPU. We use `Adagrad' optimizer with a step size of $10^{-5}$. Both training and validation accuracy reached around $90\%$ accuracy within $30$ minutes.

The dataset for the ground robot navigation (testing set) was collected at a different part of the field, marked by the red rectangle in Fig.~\ref{fig:exmap} on June 22, 2019, It contains 135 images in total. The corn rows in the testing set are planted more sparsely, as shown in Fig.~\ref{fig:contest}. There are dirt roads between rows in the testing dataset, but training set does not contain such instances. To improve the segmentation results on the testing set, we labeled three more images from the testing data which have sparse rows to fine-tune the network. Fig.~\ref{fig:fielddiff} shows an improved example after fine-tuning.
%\vtxt{I am confused by this. You use unet weights and fine tune with only three images? What happens to the remianing 97-3?}
%===============================================================================

\subsection{Simulation and Field Tests}
In this section, we investigate the performance gain attained by combining terrain class information with the onboard measurements. We conduct simulations to compare our navigation method with three benchmarks. Afterwards, we present results from field experiments.

\subsubsection{Simulations}
We designed simulation environments where the class labels and the ground truth energy costs of all the locations are available. We compare our method with three planners: $(i)$~The shortest distance planner. This planner does not have energy cost information. It returns shortest Euclidean distance path to the goal. $(ii)$~A local planner. In this method, the robot can measure the energy cost at the current location. The local planner also learns the energy cost map using Gaussian Process model. But since it does not have terrain class information, the covariance for any pair of locations is a square exponential function based on the distance, as shown in Eqn.~\ref{eq:joint}. $(iii)$~An optimal planner. It uses the ground truth cost so that an optimal path is obtained. 
%The results are shown in Table~\ref{table:cost}.

\begin{figure}
\centering
\subfigure[]{
	\includegraphics[width=.22\columnwidth]{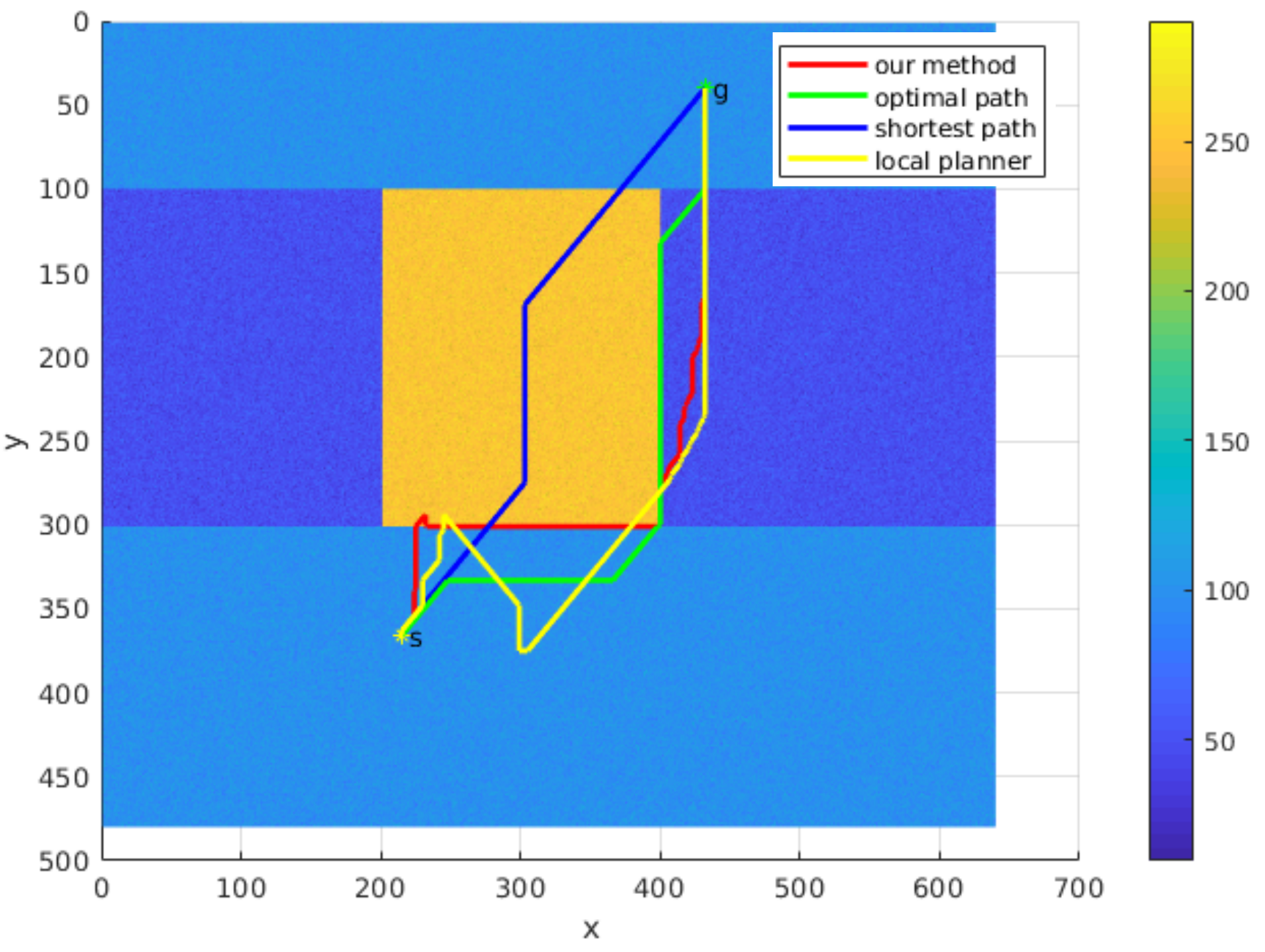}
	\label{fig:simu1}
}
\subfigure[]{	
	\includegraphics[width=.22\columnwidth]{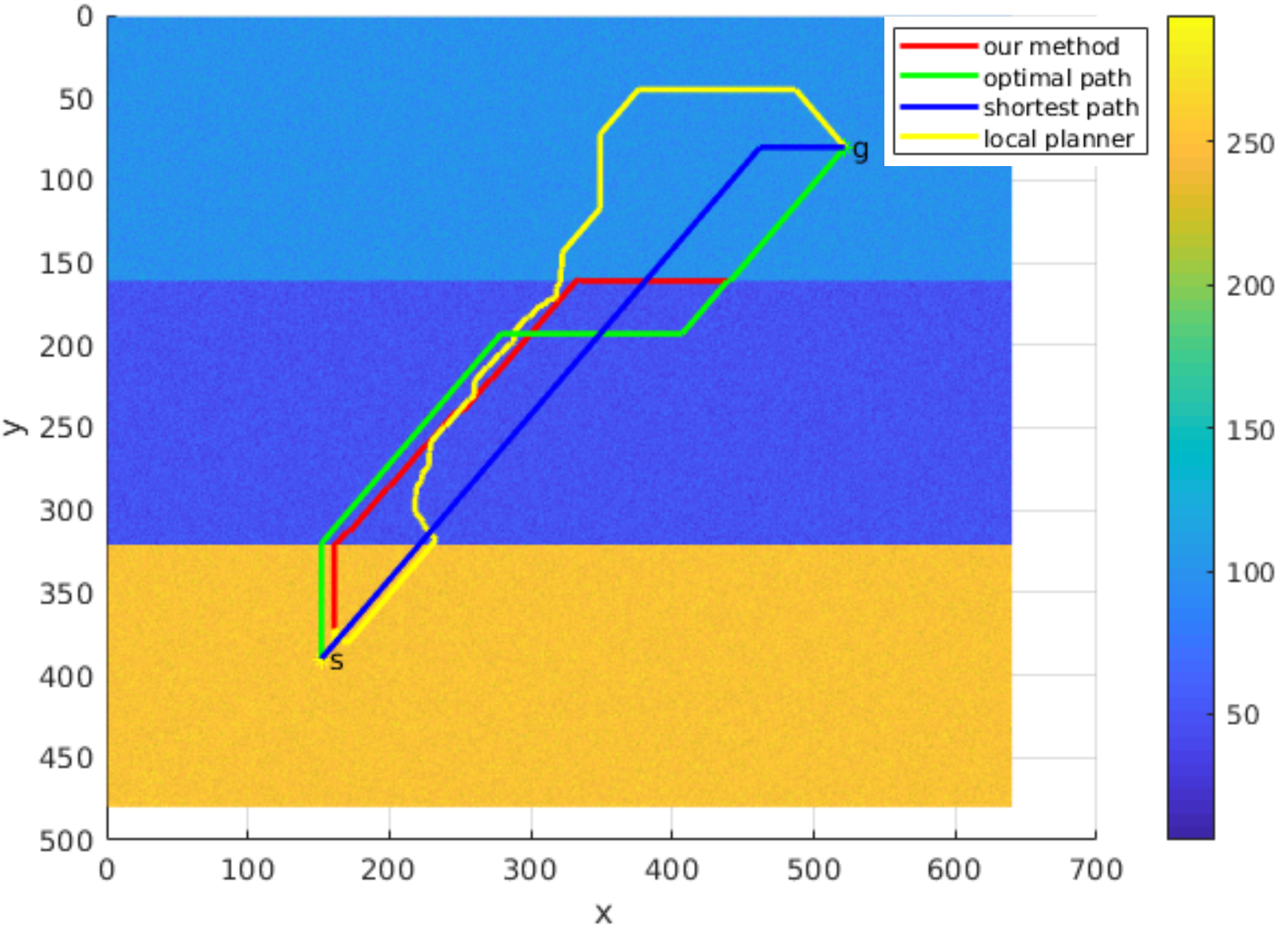}
	\label{fig:simu2}
}
\subfigure[]{
	\includegraphics[width=.23\columnwidth]{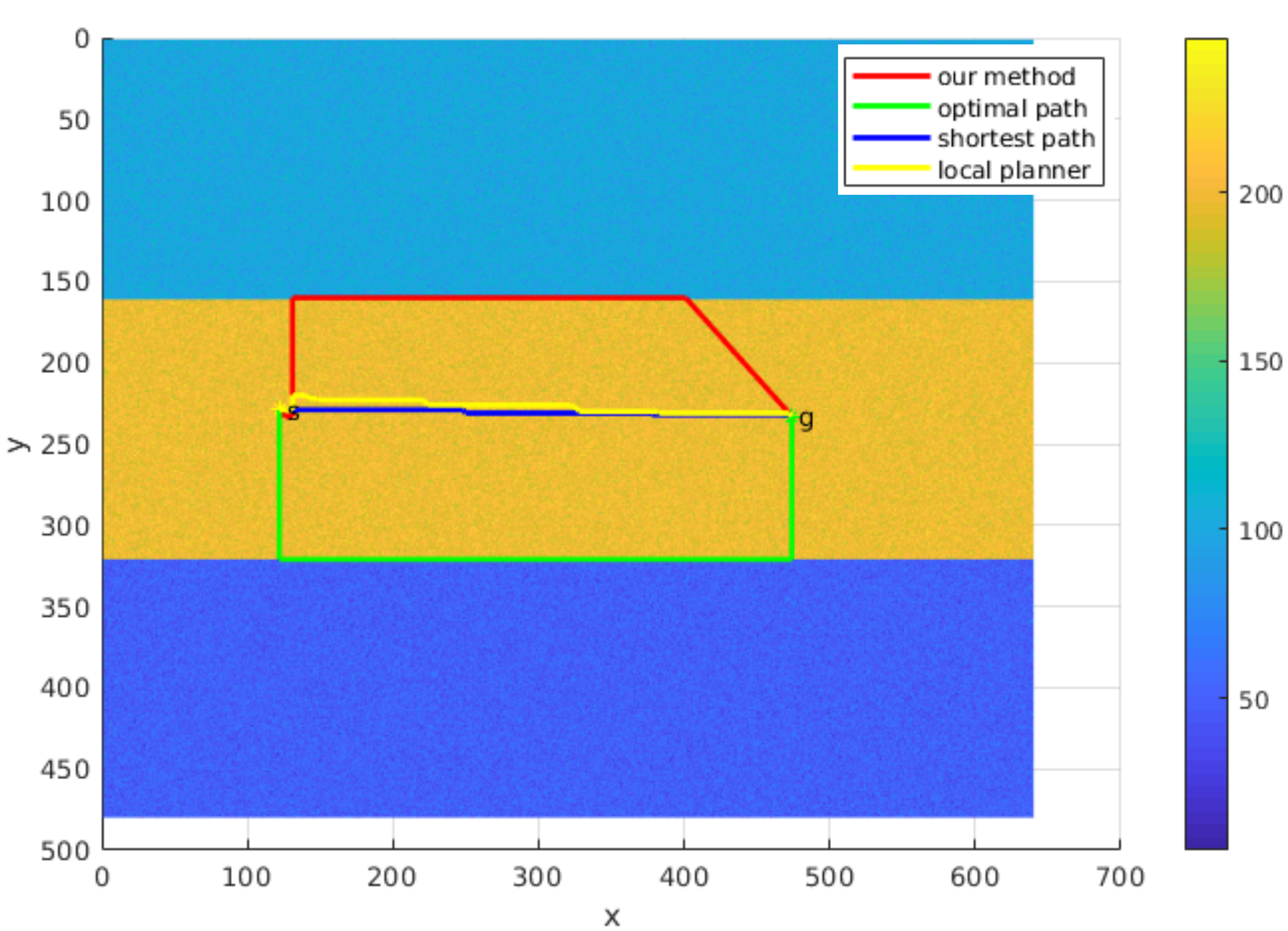}
	\label{fig:simu3}
}
\subfigure[]{	
	\includegraphics[width=.22\columnwidth]{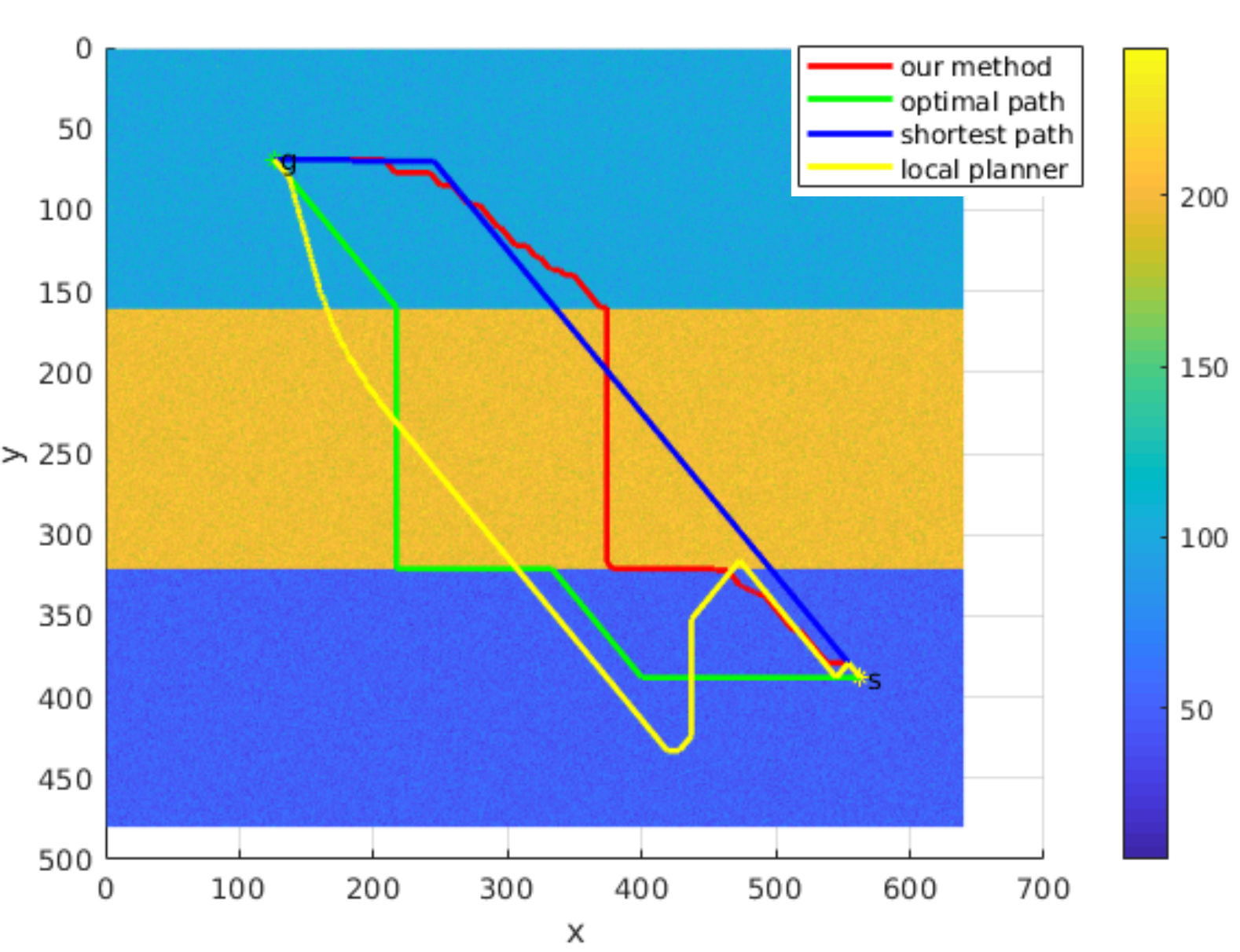}
	\label{fig:simu4}  
}
\caption{Simulation examples. The shortest Euclidean distance planner does not perform well without considering the energy cost. The local planner performs not well because it does not have terrain class information to precisely update the map. $(a),~(b)$~Our method takes a similar path to the optimal path after exploring the yellow (high cost) area. $(c)$~The terrain class at the bottom remains unexplored by our method, since navigating on the explored area takes less energy. $(d)$ The paths taken by our method and the optimal solution are different. Our method explores the yellow and light blue areas, then navigates to the goal.}
\label{fig:simuexample}
\end{figure}

Fig.~\ref{fig:simuexample} shows example runs from our simulations. The pixel colors represent the energy costs at the corresponding locations. As expected, the shortest Euclidean distance planner does not have good performance since it does not incorporate energy cost. The local planner lacks the class information to update the map precisely. For example, in Fig.~\ref{fig:simu2}, the measured energy cost increases significantly when the robot enters from the yellow area (low cost) to the dark blue area (high cost). Updating the map using Gaussian Process model will make the energy cost increase along the navigating direction. Thus the robot starts to navigate along another direction. Our method behaves similarly to the optimal solution after exploring a previously unmeasured class in the cases such as Fig.~\ref{fig:simu1} and Fig.~\ref{fig:simu2}. The map can be updated correctly with the terrain class information.

Since our method does not have complete global information, it may not always explore the lowest cost terrain (the darker blue area) first, as shown in Fig.~\ref{fig:simu3} and Fig.~\ref{fig:simu4}. In Fig.~\ref{fig:simu3}, it leaves the third area unexplored since navigating on explored terrains gives the lowest cost path. Overall our method outperforms the local planner and the shortest distance planner. Table.~\ref{table:cost} compares the energy cost ratios of different methods to the optimal solution. The energy cost difference between our solution and the optimal solution will depend on the terrain differences. In these examples, the deviation from our method to the optimal solution remains within $20\%$. 

%\vtxt{Minghan, in addition to the total energy consumption let's put (method/opt) ratios in paranthesis next to the actual numbers. For example 44364 (117\%)} done.
\begin{table}
\centering
\label{table:cost}
\begin{tabular}{|l|l|l|l|l|}
\hline
Index & Our method & Shortest distance planner & Local planner & Optimal planner \\
\hline
1 & $44364 (117\%)$ & $77641 (205\%)$ & $54972 (145\%)$ & $37836 (100\%)$ \\
\hline
2 & $47142 (101\%)$ & $53164 (114\%)$ & $70700 (152\%)$ & $46563 (100\%)$ \\
\hline
3 & $64279 (120\%)$ & $70731 (132\%)$ & $73277 (137\%)$ & $53450 (100\%)$\\
\hline
4 & $71638 (111\%)$ & $74803(116\%)$ & $78004(121\%)$ & $64279 (100\%)$ \\
\hline
\end{tabular}
\caption{Comparison of our method with three baseline approaches. The numbers in the parenthesises show the percent deviation from the optimal solution obtained using ground truth.}
\end{table}

\subsubsection{Field Tests}
\begin{figure}
\centering
\subfigure[]{
	\includegraphics[width=.24\columnwidth]{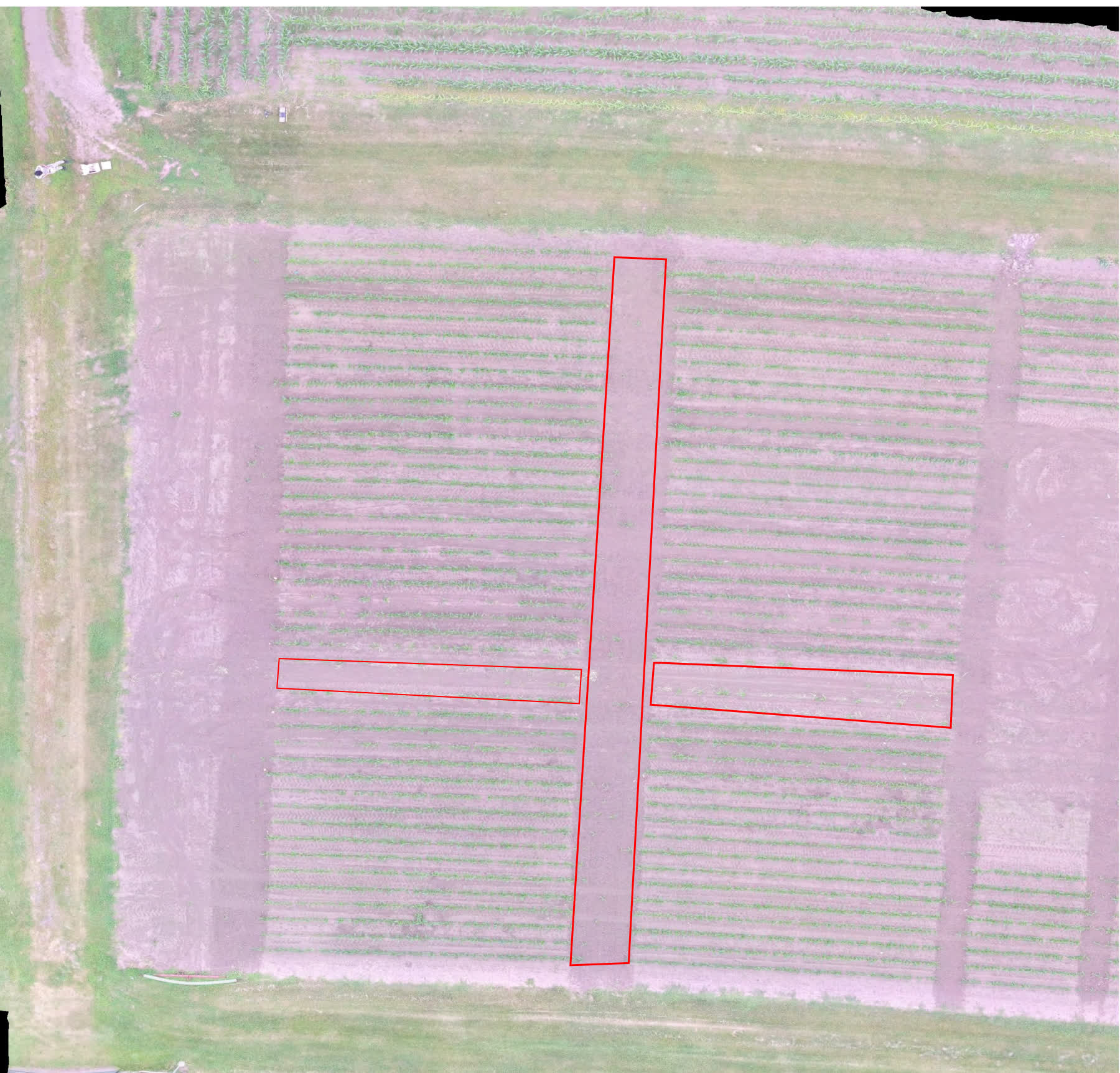}
	\label{fig:exfield}
}
\subfigure[]{	
	\includegraphics[width=.23\columnwidth]{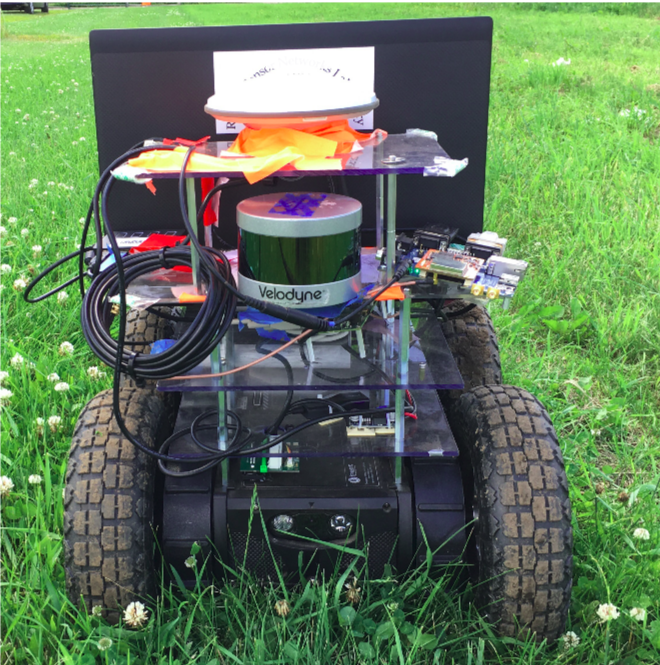}
	\label{fig:robotf}  
}
\subfigure[]{	
	\includegraphics[width=.22\columnwidth]{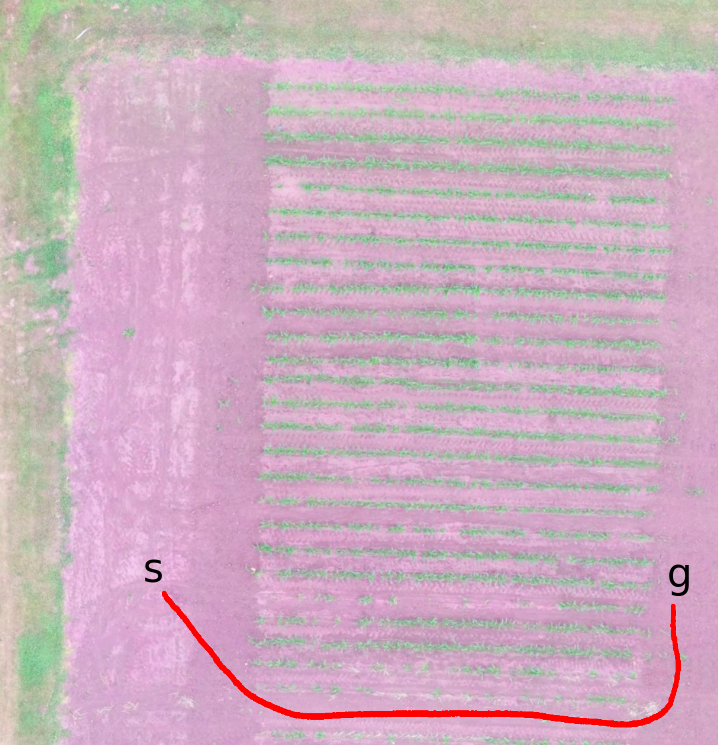}
	\label{fig:down}  
}
\subfigure[]{
	\includegraphics[width=.22\columnwidth]{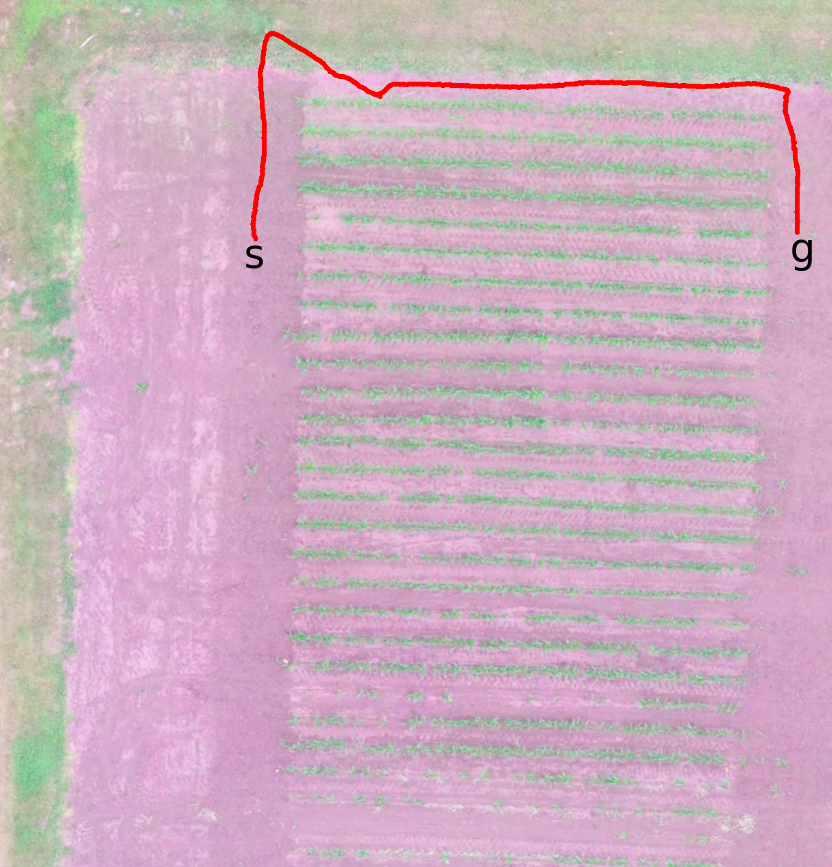}
	\label{fig:up}  
}
\caption{$(a)$ The experiment location. The red rectangles mark the wide, traversable roads. $(b)$ The ground robot for the experiment. $(c)~(d)$ With the terrain class information, the robot is able to find the efficient directions to cross the corn rows.}
\label{fig:fieldrobot}
\end{figure}
We report results from a field test in a farm setting. Fig.~\ref{fig:exfield} shows a mosaic of the aerial images obtained over the field location. The platform ``4WD rover" from Rover Robotics as shown in Fig.~\ref{fig:robotf}, was used in the experiments. The robot is equipped with an RTK GPS to get position data with centimeter-level accuracy. The onboard sensors, including the laptop, are powered by separate batteries. The energy consumption of the robot battery is mainly due to the motion. For safety, we require the robot to stay clear of the plants. Thus in our experiment, only the wide dirt roads between groups of corn rows are traversable. These areas are marked with red rectangles in Fig.~\ref{fig:exfield}.

Fig.~\ref{fig:down} and Fig.~\ref{fig:up} demonstrate examples in which, with global terrain information, the robot is able to find the efficient directions (short paths) to cross the corn rows. It is easy to see that if the robot only has local energy consumption measurements, it may not be able to follow the shorter paths to cross the corn rows to the goal positions.

\begin{figure}
\centering
\subfigure[]{	
	\includegraphics[width=.265\columnwidth]{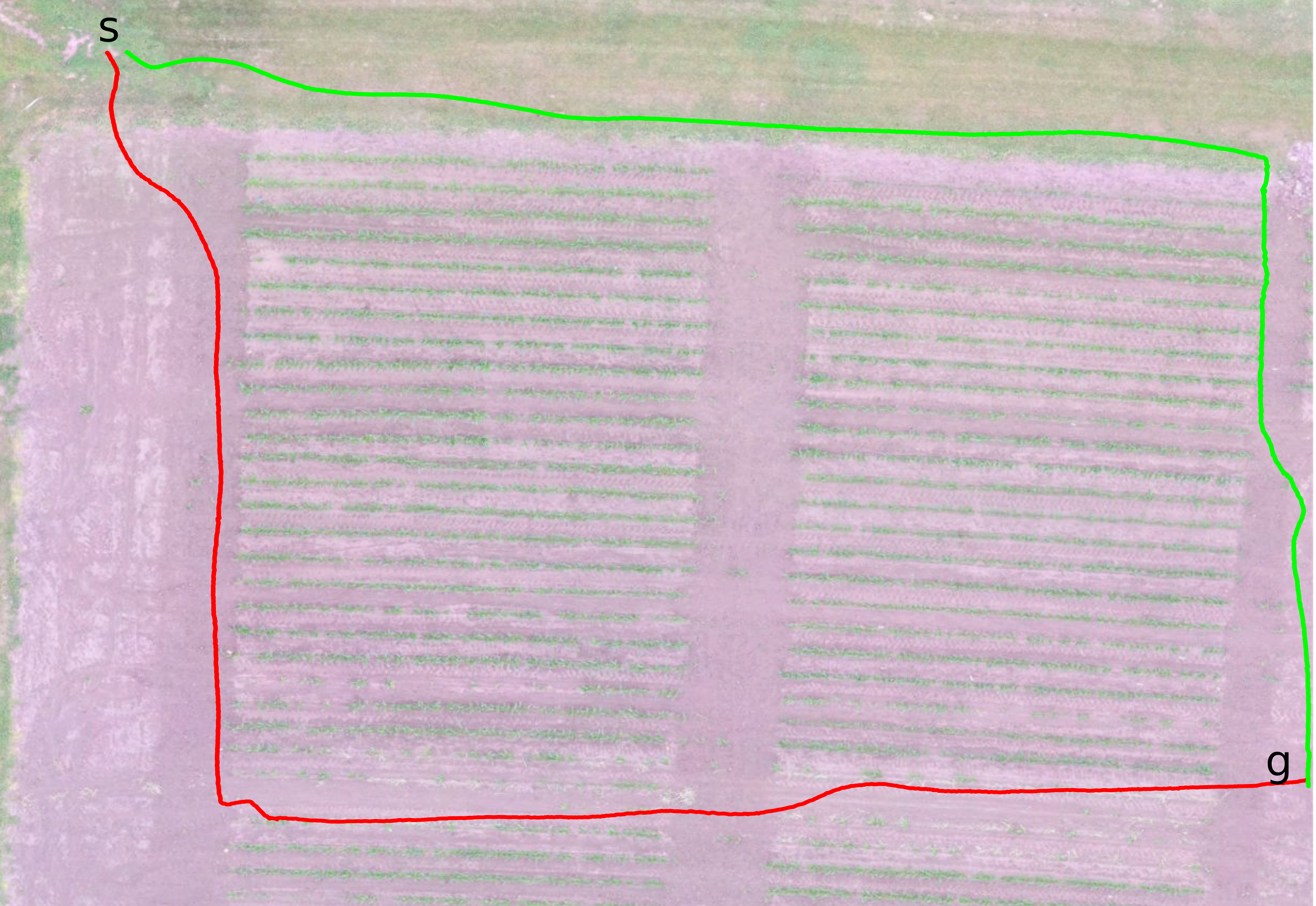}
	\label{fig:p1image}  
}
\subfigure[]{
	\includegraphics[width=.235\columnwidth]{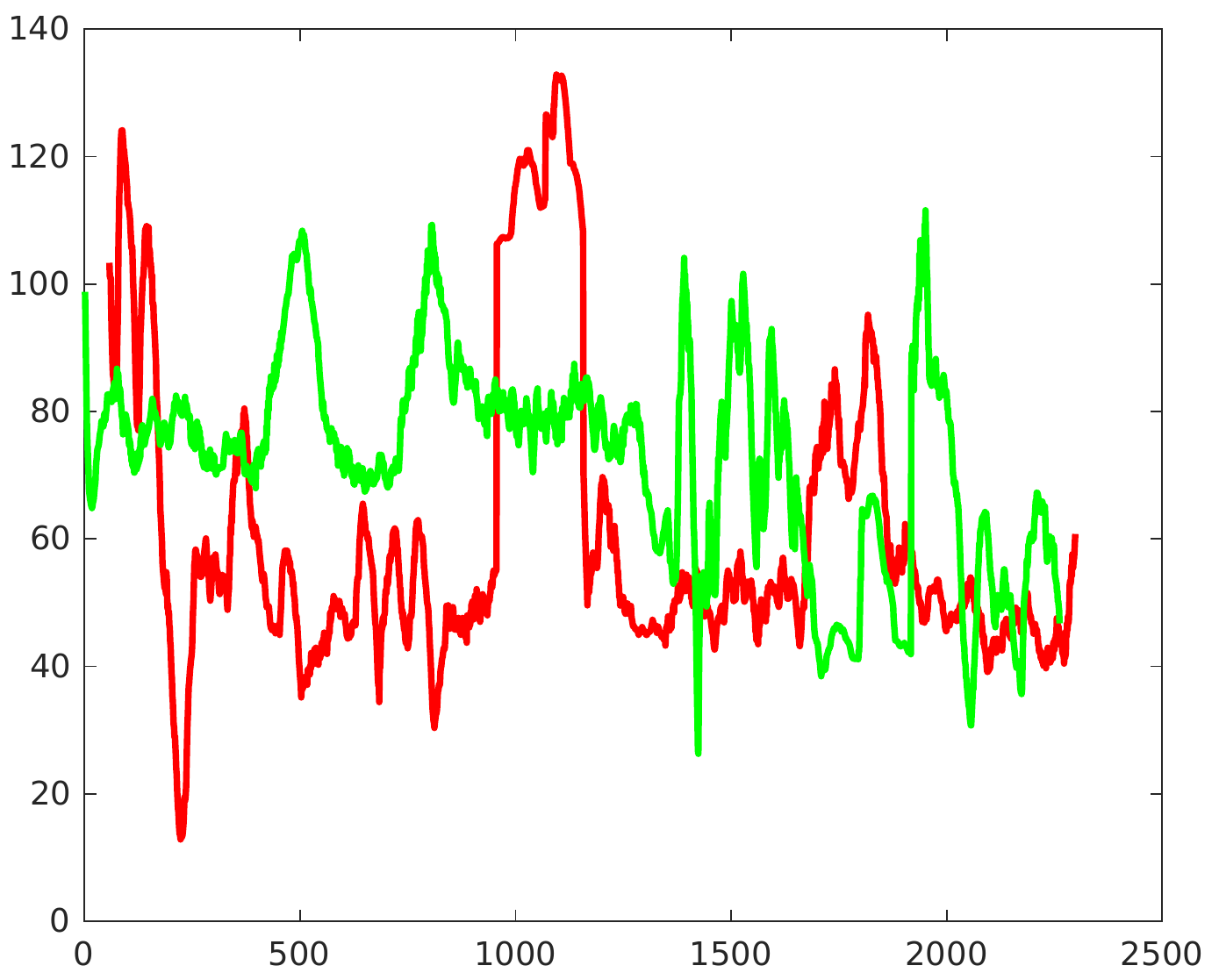}
	\label{fig:p1energy}  
}
\subfigure[]{	
	\includegraphics[width=.145\columnwidth]{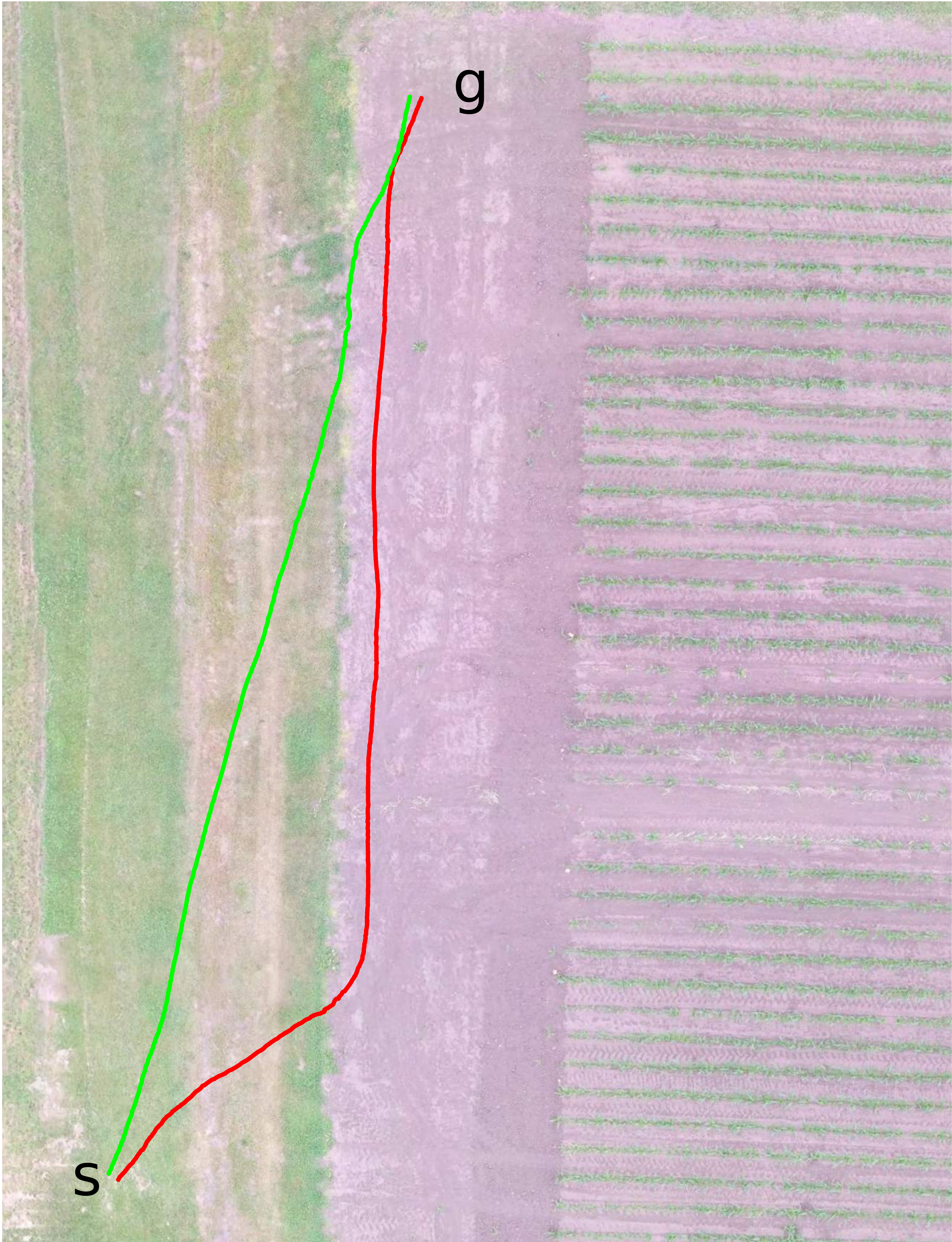}
	\label{fig:p2image}  
}
\subfigure[]{	
	\includegraphics[width=.255\columnwidth]{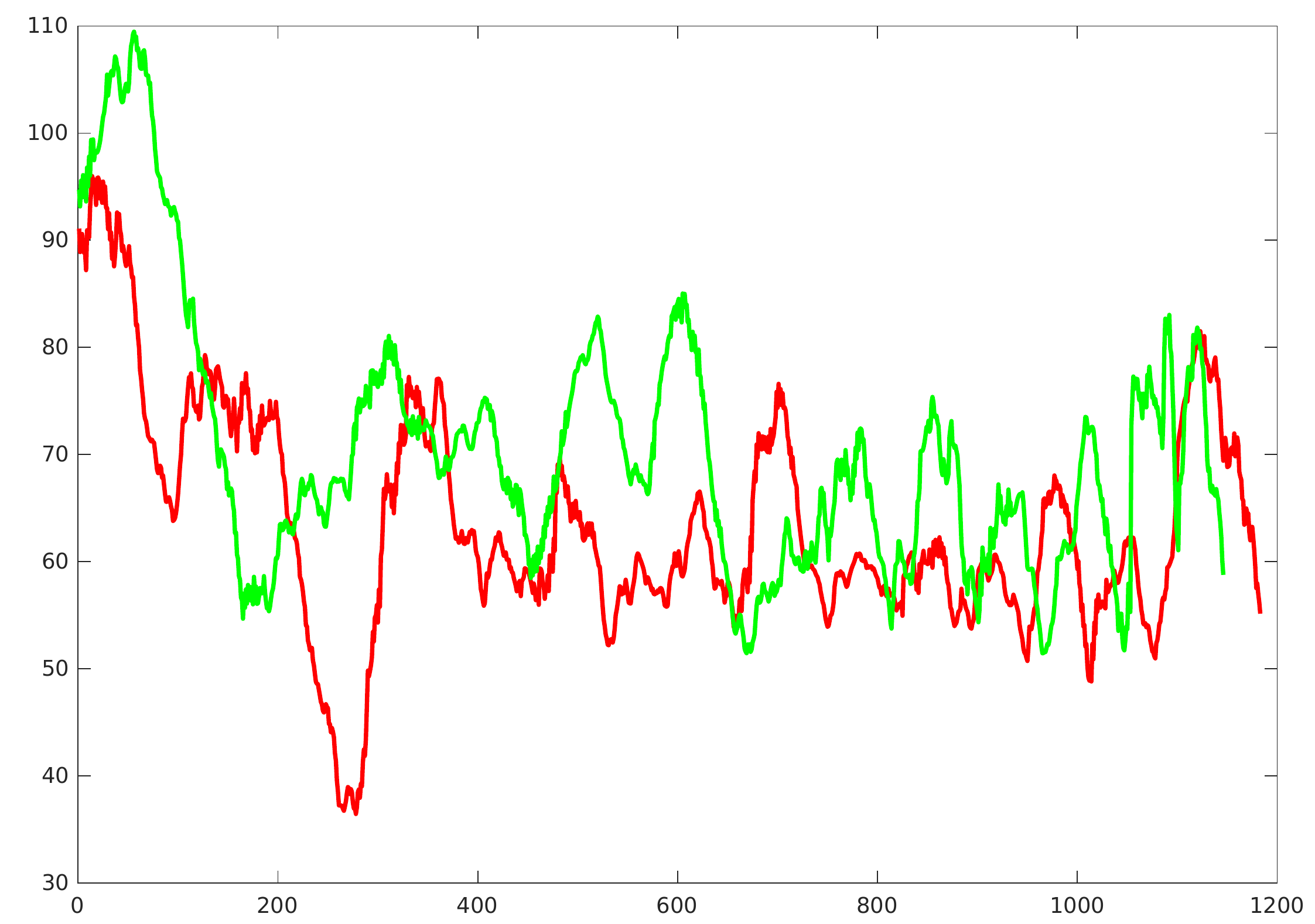}
	\label{fig:p2energy}  
}
\caption{Comparing the path planned by our method (red) with shorter paths (green) from $s$ to $g$. $(a)~(c)$~The trajectories of the robot plotted on the images. $(b)~(d)$~The unit-distance energy cost along the trajectories (unit: J/m).}
\label{fig:p1}
\end{figure}

Next, we show in Fig.~\ref{fig:p1} that our method can not only find short paths, but also energy-efficient paths. The shortest paths from the starting positions to the goals are along the green trajectories. Using our method, the robot explores the dirt area first. It finds that the dirt area has a smaller unit-distance energy cost. Thus the robot takes the longer path on the dirt. Fig.~\ref{fig:p1energy} and Fig.~\ref{fig:p2energy} show the unit-distance energy cost measurements along the trajectories in Fig.~\ref{fig:p1image} and Fig.~\ref{fig:p2image}. Note that there is a height gap on the boundary between the dirt area and the grass. Meanwhile, turning usually costs more energy than moving forward. Thus we see some peaks/valleys in Fig.~\ref{fig:p1energy} and Fig.~\ref{fig:p2energy}.

In Fig.~\ref{fig:p1image}, the length of the green path is $54.9m$. The robot consumes around $4647.9J$ energy to follow it. The length of the red trajectory is $58.5m$, while it takes $3729.3J$ energy. It saves $19.7\%$ energy compared to the shortest path. In Fig.~\ref{fig:p2image}, the length of the green trajectory is $26.7m$, and it takes $1988.4J$ energy. The length of the red trajectory is $28.5m$, and it consumes $1751.5J$ energy. In this test our method saves around $11.9\%$ energy. 

Fig.~\ref{fig:up} also identifies a limitation inherent in online approaches: Without accurate prior information, the robot will spend additional cost to explore other terrain classes if the current estimation gives a potential better path on these classes.

\section{Conclusion}
\label{sec:conclusion}
In this paper, we presented an energy-efficient path planning method. By combining aerial images and robot onboard measurements, we performed energy cost map estimation and path planning simultaneously. The novel aspect of our algorithm is that a covariance function based on terrain classification yields more effective navigation strategies as compared to standard approaches. The experiments in an agriculture setting show that our method finds more energy-efficient paths than those given by distance-based shortest paths. 

Our method starts with a segmented image to provide class information. In our future work, we will investigate the sensitivity of our method to segmentation accuracy. We will also incorporate more general terrain information such as slope in the energy cost map. Finally, we would like to extend the approach to multiple robots.

%===============================================================================

% The maximum paper length is 8 pages excluding references and acknowledgements, and 10 pages including references and acknowledgements

\clearpage
% The acknowledgments are automatically included only in the final version of the paper.
\acknowledgments{This work is funded in part by NSF grant \# 1525045, NSF grant \#1617718, and MN State LCCMR program.}

%===============================================================================

% no \bibliographystyle is required, since the corl style is automatically used.
\bibliography{reference.bib}  % .bib

\begin{thebibliography}{21}
\providecommand{\natexlab}[1]{#1}
\providecommand{\url}[1]{\texttt{#1}}
\expandafter\ifx\csname urlstyle\endcsname\relax
  \providecommand{\doi}[1]{doi: #1}\else
  \providecommand{\doi}{doi: \begingroup \urlstyle{rm}\Url}\fi

\bibitem[Sun and Reif(2005)]{sun2005finding}
Z.~Sun and J.~H. Reif.
\newblock On finding energy-minimizing paths on terrains.
\newblock \emph{IEEE Transactions on Robotics}, 21\penalty0 (1):\penalty0
  102--114, 2005.

\bibitem[Salan et~al.(2014)Salan, Drumwright, and Lin]{salan2014minimum}
S.~Salan, E.~Drumwright, and K.-I. Lin.
\newblock Minimum-energy robotic exploration: A formulation and an approach.
\newblock \emph{IEEE Transactions on Systems, Man, and Cybernetics: Systems},
  45\penalty0 (1):\penalty0 175--182, 2014.

\bibitem[Ersson and Hu()]{ersson2001path}
T.~Ersson and X.~Hu.
\newblock Path planning and navigation of mobile robots in unknown
  environments.
\newblock In \emph{Proceedings 2001 IEEE/RSJ International Conference on
  Intelligent Robots and Systems. Expanding the Societal Role of Robotics in
  the the Next Millennium (Cat. No. 01CH37180)}, volume~2, pages 858--864.
  IEEE.

\bibitem[LaValle(1998)]{lavalle1998rapidly}
S.~M. LaValle.
\newblock Rapidly-exploring random trees: A new tool for path planning.
\newblock 1998.

\bibitem[Kuffner~Jr and LaValle(2000)]{kuffner2000rrt}
J.~J. Kuffner~Jr and S.~M. LaValle.
\newblock Rrt-connect: An efficient approach to single-query path planning.
\newblock In \emph{ICRA}, volume~2, 2000.

\bibitem[Atramentov and LaValle(2002)]{atramentov2002efficient}
A.~Atramentov and S.~M. LaValle.
\newblock Efficient nearest neighbor searching for motion planning.
\newblock In \emph{Proceedings 2002 IEEE International Conference on Robotics
  and Automation (Cat. No. 02CH37292)}, volume~1, pages 632--637. IEEE, 2002.

\bibitem[Koenig and Likhachev(2005)]{koenig2005fast}
S.~Koenig and M.~Likhachev.
\newblock Fast replanning for navigation in unknown terrain.
\newblock \emph{IEEE Transactions on Robotics}, 21\penalty0 (3):\penalty0
  354--363, 2005.

\bibitem[Howard and Kelly(2007)]{howard2007optimal}
T.~M. Howard and A.~Kelly.
\newblock Optimal rough terrain trajectory generation for wheeled mobile
  robots.
\newblock \emph{The International Journal of Robotics Research}, 26\penalty0
  (2):\penalty0 141--166, 2007.

\bibitem[Paden et~al.(2016)Paden, {\v{C}}{\'a}p, Yong, Yershov, and
  Frazzoli]{paden2016survey}
B.~Paden, M.~{\v{C}}{\'a}p, S.~Z. Yong, D.~Yershov, and E.~Frazzoli.
\newblock A survey of motion planning and control techniques for self-driving
  urban vehicles.
\newblock \emph{IEEE Transactions on intelligent vehicles}, 1\penalty0
  (1):\penalty0 33--55, 2016.

\bibitem[Tiwari et~al.(2018)Tiwari, Xiao, and Chong]{tiwari2018estimating}
K.~Tiwari, X.~Xiao, and N.~Y. Chong.
\newblock Estimating achievable range of ground robots operating on single
  battery discharge for operational efficacy amelioration.
\newblock In \emph{2018 IEEE/RSJ International Conference on Intelligent Robots
  and Systems (IROS)}, pages 3991--3998. IEEE, 2018.

\bibitem[Liu and Sun(2014)]{liu2014minimizing}
S.~Liu and D.~Sun.
\newblock Minimizing energy consumption of wheeled mobile robots via optimal
  motion planning.
\newblock \emph{IEEE/ASME Transactions on Mechatronics}, 19\penalty0
  (2):\penalty0 401--411, 2014.

\bibitem[Ganganath et~al.(2015)Ganganath, Cheng, and
  Chi]{ganganath2015constraint}
N.~Ganganath, C.-T. Cheng, and K.~T. Chi.
\newblock A constraint-aware heuristic path planner for finding
  energy-efficient paths on uneven terrains.
\newblock \emph{IEEE transactions on industrial informatics}, 11\penalty0
  (3):\penalty0 601--611, 2015.

\bibitem[Schilling et~al.(2017)Schilling, Chen, Folkesson, and
  Jensfelt]{schilling2017geometric}
F.~Schilling, X.~Chen, J.~Folkesson, and P.~Jensfelt.
\newblock Geometric and visual terrain classification for autonomous mobile
  navigation.
\newblock In \emph{2017 IEEE/RSJ International Conference on Intelligent Robots
  and Systems (IROS)}, pages 2678--2684. IEEE, 2017.

\bibitem[Chavez-Garcia et~al.(2018)Chavez-Garcia, Guzzi, Gambardella, and
  Giusti]{chavez2018learning}
R.~O. Chavez-Garcia, J.~Guzzi, L.~M. Gambardella, and A.~Giusti.
\newblock Learning ground traversability from simulations.
\newblock \emph{IEEE Robotics and Automation Letters}, 3\penalty0 (3):\penalty0
  1695--1702, 2018.

\bibitem[Mei et~al.(2004)Mei, Lu, Hu, and Lee]{mei2004energy}
Y.~Mei, Y.-H. Lu, Y.~C. Hu, and C.~G. Lee.
\newblock Energy-efficient motion planning for mobile robots.
\newblock In \emph{IEEE International Conference on Robotics and Automation,
  2004. Proceedings. ICRA'04. 2004}, volume~5, pages 4344--4349. IEEE, 2004.

\bibitem[Tokekar et~al.(2014)Tokekar, Karnad, and Isler]{tokekar2014energy}
P.~Tokekar, N.~Karnad, and V.~Isler.
\newblock Energy-optimal trajectory planning for car-like robots.
\newblock \emph{Autonomous Robots}, 37\penalty0 (3):\penalty0 279--300, 2014.

\bibitem[Mei et~al.(2006)Mei, Lu, Lee, and Hu]{mei2006energy}
Y.~Mei, Y.-H. Lu, C.~G. Lee, and Y.~C. Hu.
\newblock Energy-efficient mobile robot exploration.
\newblock In \emph{Proceedings 2006 IEEE International Conference on Robotics
  and Automation, 2006. ICRA 2006.}, pages 505--511. IEEE, 2006.

\bibitem[Chekuri and Pal(2005)]{chekuri2005recursive}
C.~Chekuri and M.~Pal.
\newblock A recursive greedy algorithm for walks in directed graphs.
\newblock In \emph{46th Annual IEEE Symposium on Foundations of Computer
  Science (FOCS'05)}, pages 245--253. IEEE, 2005.

\bibitem[Singh et~al.(2009)Singh, Krause, Guestrin, and
  Kaiser]{singh2009efficient}
A.~Singh, A.~Krause, C.~Guestrin, and W.~J. Kaiser.
\newblock Efficient informative sensing using multiple robots.
\newblock \emph{Journal of Artificial Intelligence Research}, 34:\penalty0
  707--755, 2009.

\bibitem[Rasmussen and Nickisch(2010)]{rasmussen2010gaussian}
C.~E. Rasmussen and H.~Nickisch.
\newblock Gaussian processes for machine learning (gpml) toolbox.
\newblock \emph{Journal of machine learning research}, 11\penalty0
  (Nov):\penalty0 3011--3015, 2010.

\bibitem[Ronneberger et~al.(2015)Ronneberger, Fischer, and
  Brox]{ronneberger2015u}
O.~Ronneberger, P.~Fischer, and T.~Brox.
\newblock U-net: Convolutional networks for biomedical image segmentation.
\newblock In \emph{International Conference on Medical image computing and
  computer-assisted intervention}, pages 234--241. Springer, 2015.

\end{thebibliography}

\end{document}